\documentclass[runningheads]{llncs}

\usepackage{eccv}

\usepackage{eccvabbrv}

\usepackage{graphicx}
\usepackage{booktabs}
\usepackage{wrapfig}
\usepackage{makecell}

\usepackage[accsupp]{axessibility}  

\usepackage{hyperref}

\usepackage{orcidlink}

\usepackage{amsmath,amsfonts,bm}

\def\eqref#1{equation~\ref{#1}}

\def\1{\bm{1}}

\DeclareMathAlphabet{\mathsfit}{\encodingdefault}{\sfdefault}{m}{sl}
\SetMathAlphabet{\mathsfit}{bold}{\encodingdefault}{\sfdefault}{bx}{n}

\usepackage{hyperref}
\usepackage{url}
\usepackage[utf8]{inputenc} 
\usepackage[T1]{fontenc}    
\usepackage{booktabs}       
\usepackage{nicefrac}       

\usepackage{microtype}      
\usepackage{xcolor}         
\usepackage{amsmath}
\usepackage[ruled,vlined]{algorithm2e}
\usepackage{amssymb}
\usepackage{colortbl} 
\usepackage{mwe}
\usepackage{multicol}
\usepackage{multirow}
\usepackage{caption}
\newcommand{\cmark}{\textcolor{green!60!black}{\checkmark}}%
\newcommand{\xmark}{\textcolor{red}{\sffamily X}}%
\definecolor{dodgerblue}{RGB}{28,134,238}
\definecolor{crimson}{RGB}{220,20,60}
\definecolor{slateblue}{RGB}{100,149,237}
\usepackage{amsmath,amsfonts,bm}

\newcommand{\fref}[1]{Figure~\ref{#1}}

\begin{document}

\title{JAM-Flow: Joint Audio-Motion Synthesis \\ with Flow Matching}


\author{
Mingi Kwon\textsuperscript{*}\inst{1,2}
\and
Joonghyuk Shin\textsuperscript{*}\inst{3}
\and
Jaeseok Jeong\inst{1,2}
\and
\\ Jaesik Park\textsuperscript{\dag}\inst{3}
\and
Youngjung Uh\textsuperscript{\dag}\inst{1}
}

\authorrunning{M.~Kwon et al.}


\institute{Yonsei University\\ 
\email{\{kwonmingi,jete\_jeong,yj.uh\}@yonsei.ac.kr} \and
CineLingo\and
Seoul National University\\
\email{\{joonghyuk,jaesik.park\}@snu.ac.kr}}

\maketitle
\footnotetext{*Equal Contribution, \dag Equal Advising}

\begin{abstract}
The intrinsic link between facial motion and speech is often overlooked in generative modeling, where talking head synthesis and text-to-speech (TTS) are typically addressed as separate tasks. This paper introduces JAM-Flow, a unified framework to simultaneously synthesize and condition on both facial motion and speech. Our approach leverages flow matching and a novel Multi-Modal Diffusion Transformer architecture, integrating specialized Motion-DiT and Audio-DiT modules. These are coupled via selective joint attention layers and incorporate key architectural choices, such as temporally aligned positional embeddings and localized joint attention masking, to enable effective cross-modal interaction while preserving modality-specific strengths. By analyzing and leveraging pretrained representation embeddings, JAM-Flow is designed for efficient, near real-time sampling. Trained with an inpainting-style objective, JAM-Flow supports a wide array of conditioning inputs (including text, reference audio, and reference motion) facilitating tasks such as synchronized talking head generation from text, audio-driven animation, and much more, within a single, coherent model. JAM-Flow significantly advances multi-modal generative modeling by providing a practical solution for holistic audio-visual synthesis. \href{https://joonghyuk.com/jamflow-web/}{project page}
\keywords{Joint Audio-Motion Generation \and Talking Head Generation \and Text-to-Speech}
\end{abstract}

\section{Introduction}
With the rapid advancement of generative models~\cite{goodfellow2014generative, ho2020denoising, song2021scorebased, liuflow, lipmanflow, rombach2022high}, the synthesis of realistic human faces and voices has become increasingly sophisticated. Two major fields have emerged from this trend: talking head generation~\cite{guo2024liveportrait, xie2024x, siarohin2019first, drobyshev2024emoportraits, wang2021one, zhou2020makeittalk, tian2024emo, xu2024vasa, lin2025omnihuman}, which animates static portrait images to mimic facial expressions, and text-to-speech (TTS) synthesis~\cite{chen2024f5, jiang2025megatts, eskimez2024e2, guo2024fireredtts, du2024cosyvoice}, which converts text and a short voice reference into natural-sounding speech. While state-of-the-art methods in both domains, ranging from GAN-based models~\cite{guo2024liveportrait, zhang2023sadtalker} for near real-time inference to diffusion-based models~\cite{xie2024x, lin2025omnihuman} and flow matching-based models~\cite{chen2024f5, jiang2025megatts} for higher fidelity, have made remarkable progress, these two problems have traditionally been treated as separate tasks.

\begin{figure}[h]
  \centering
  \includegraphics[width=1.0\textwidth]{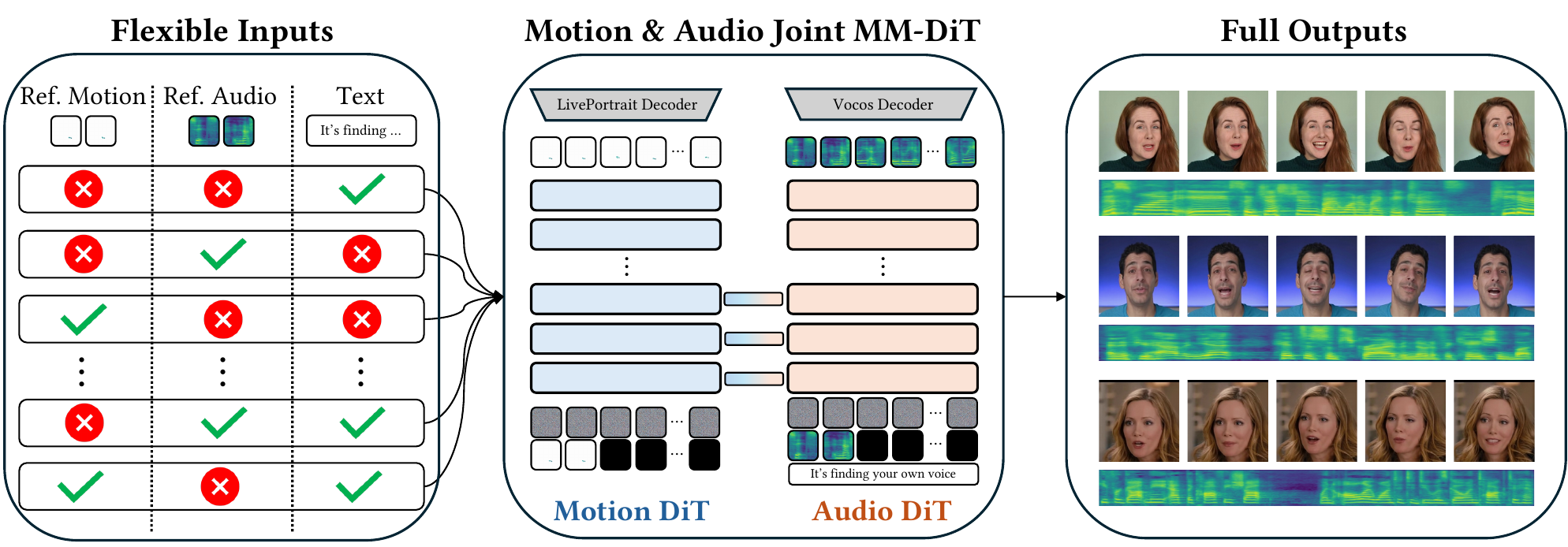}
  \caption{Overview of our JAM-Flow framework for flexible and joint generation of facial motion and speech. The model accepts diverse input combinations, including text, reference motion, and reference audio. These are processed by our novel Motion \& Audio Joint MM-DiT, which enables synchronized synthesis of full audio-visual outputs supporting tasks like talking head generation from text, audio-driven animation, and cross-modal reconstruction (e.g., audio from motion).}
  \label{fig:teaser}
\end{figure}

Yet in natural human communication, not only is real-time-like latency important, but facial motion and speech are also deeply interwoven. Movements of the mouth, cheeks, and jaw are not merely visual artifacts but integral components of spoken language. Surprisingly, despite this intrinsic connection, few prior works have jointly addressed talking head generation and speech synthesis in a unified model. Existing talking head models typically treat audio as a unidirectional condition, while TTS systems remain blind to facial dynamics.

In this work, we introduce the first training framework that can simultaneously model, generate, and condition on both audio and facial motion modalities within a single flow matching-based framework, while supporting generation at near real-time latency.
We integrate two specialized flow matching models: a Motion-DiT for generating implicit facial keypoint sequences, and an Audio-DiT for denoising mel-spectrograms from text and reference speech. These modules are coupled through selective joint attention layers, where only half the layers are fused, allowing effective cross-modal communication while preserving the benefits of modality-specific representations. Furthermore, to inject structural inductive biases into the model, we incorporate rotary positional embeddings (RoPE)~\cite{su2024roformer} with task-specific engineering improvements and attention masking to restrict temporal receptive fields.

One of our key experimental observations is that models should be initialized from well-trained components in their respective modalities before learning cross-modal relations. In our framework, we initialize the TTS branch with a pretrained F5-TTS~\cite{chen2024f5} model, while the Motion-DiT is first trained separately using the LivePortrait~\cite{guo2024liveportrait} framework to capture facial motion via compact keypoints. After this modality-specific pretraining, the two modules are jointly trained with shared attention layers and an inpainting-style supervision scheme. We propose that this joint inpainting-style supervision is particularly well-suited for optimizing pretrained unimodal models to learn robust cross-modal interactions, enabling flexible generation even under partially missing inputs.

Our joint inpainting-style supervision not only facilitates effective cross-modal learning but also enables versatile inference. By design, each modality can be conditioned on full, partial, or even absent inputs, allowing a single model to flexibly support diverse tasks and generation scenarios.
As illustrated in \fref{fig:teaser}, our model supports a wide range of input-output configurations, including talking head generation, TTS, and cross-modal reconstruction, within a single coherent framework.
To the best of our knowledge, this is the first attempt to employ joint inpainting-style supervision to achieve such flexible cross-modal synthesis, marking an important step toward practical and scalable multimodal generation.

Our contributions are summarized as follows:
\begin{enumerate}
\item We analyze and leverage pretrained implicit representations to enable talking head video generation and TTS synthesis at near real-time latency.
\item We present the first inpainting-style joint training framework for talking head and TTS generation, which integrates both the model architecture and training strategy to enable mutual conditioning across modalities.
\item We demonstrate that our inpainting-style joint supervision effectively learns cross-modal relations, allowing robust and versatile generation across diverse tasks.
\item We further provide a practical architectural design for connecting pretrained unimodal models, including partial joint attention, RoPE integration, and attention masking, tailored for multi-modal flow matching.
\end{enumerate}

\section{Related Work}
\subsection{Flow Matching and Multimodal Diffusion Transformer}
Flow matching~\cite{liuflow, lipmanflow, esser2024scaling} enhances generative modeling efficiency over score-based diffusion models~\cite{ho2020denoising, song2021scorebased}. It learns a continuous transformation $Z_t$ between distributions via an Ordinary Differential Equation (ODE): $dZ_t/dt = v_\theta(Z_t, t)$, where $v_\theta$ is a learnable vector field. The objective is to match $v_\theta$ to a target velocity field $u_t(x)$.
Conditional Flow Matching (CFM)~\cite{lipmanflow} and Rectified Flow (RF)~\cite{liuflow} simplify this: for paired samples $x_0 \sim \pi_0$ and $x_1 \sim \pi_1$, they define an intermediate point via linear interpolation, $x_t = (1-t)x_0 + tx_1$, and set the target velocity $u_t(x_t)$ to be the difference $x_1 - x_0$. This results in the CFM loss:
\begin{equation}
\mathcal{L}_{\text{CFM}}(\theta) = \mathbb{E}_{t, x_0, x_1} \left[\|v_\theta(x_t, t) - (x_1 - x_0)\|^2\right].
\label{eq:l_cfm}
\end{equation}

The \textit{Multimodal Diffusion Transformer} (MM-DiT)~\cite{esser2024scaling} builds on flow matching for joint multi-modal generation. It uses separate DiT~\cite{peebles2023scalable} branches per modality, fused via joint self-attention for cross-modal interaction, enabling scalable and expressive generation. Large-scale MM-DiTs like Stable Diffusion 3~\cite{esser2024scaling}, Flux~\cite{flux2024}, and CogVideoX~\cite{yang2024cogvideox} achieve SOTA in various tasks (e.g., text-to-image, video synthesis), but none have explored inpainting-style joint training to simultaneously synthesize two modalities.

\subsection{Talking Head Generation}
Talking head generation synthesizes realistic facial animations from conditional signals, mainly through video-driven (facial reenactment) or audio-driven methods. Video-driven approaches~\cite{guo2024liveportrait, xie2024x, siarohin2019first} animate a source portrait using motion cues from a driving video. Early methods often used facial landmarks~\cite{siarohin2019first} or 3D models~\cite{tewari2020stylerig}, while recent works like LivePortrait~\cite{guo2024liveportrait} use implicit keypoints and X-portrait~\cite{xie2024x} employs diffusion control. In contrast, audio-driven methods~\cite{prajwal2020wav2lip, zhou2020makeittalk, tian2024emo, xu2024vasa} create lip movements synchronized with input audio. Pioneering work like Wav2Lip~\cite{prajwal2020wav2lip} focused on lip-sync accuracy with GAN, while later methods (e.g., EMO~\cite{tian2024emo}, Omni-Human~\cite{lin2025omnihuman}) often use diffusion models for enhanced expressiveness.

Current methods~\cite{gao2025wan, ma2025playmate2, chen2025humo, tu2025stableavatar, gan2025omniavatar, chen2025hunyuanvideo, yang2025infinitetalk, wang2025talkverse, seo2025lookahead, li2025infinityhuman, jiang2025omnihuman, tian2025emo2, guan2025audcast, wang2026joyavatar, zhang2025soul, ding2025kling} typically model a uni-directional audio-to-visual flow based on pre-trained large-scale text-to-video diffusion transformers. However, facial motion and speech are mutually influential in real conversations. Our hybrid approach addresses this by combining joint diffusion-based generation of keypoints and audio with efficient pixel decoding via LivePortrait, enabling bidirectional information exchange and flexible generation.

\subsection{Neural Text-to-Speech Generation}

Neural text-to-speech (TTS) has progressed from attention-based sequence-to-sequence models~\cite{wang2017tacotron} to more advanced architectures. Early non-autoregressive systems~\cite{ren2019fastspeech, ren2020fastspeech} enhanced efficiency via explicit duration modeling. A significant shift towards zero-shot voice cloning was pioneered by neural codec language models~\cite{wang2023neural}, which autoregressively generate speech tokens from minimal reference audio, despite some inference challenges.

Diffusion-based methods have since emerged as a powerful paradigm, particularly for zero-shot voice cloning. These include approaches demonstrating high-quality speech via diffusion~\cite{tan2024naturalspeech, ju2024naturalspeech, shen2023naturalspeech, le2023voicebox} and flow matching for efficient text-guided speech infilling~\cite{chen2024f5}. Recent efforts focus on refining speech-text alignment and moving from explicit duration to more flexible frameworks~\cite{lee2024ditto, eskimez2024e2}. F5-TTS~\cite{chen2024f5} advances this with flow matching and Diffusion Transformers (DiTs) for efficient non-autoregressive generation. While research continues to address alignment robustness, prosody, and efficiency~\cite{jiang2025megatts}, the simultaneous generation of speech and matching lip motion remains largely unaddressed.

\subsection{Automated Video Dubbing}
Automated video dubbing synthesizes speech from text and video inputs, focusing on aligning generated speech with existing visual content, unlike video generation from audio/text. It extends text-to-speech (TTS) by conditioning on visual context, especially lip movements and facial expressions. The main challenge is ensuring temporal synchronization and reflecting visual expressiveness in the synthesized speech. Prior work~\cite{cong2023learning, cong2024styledubber, sung2025voicecraft} has explored aligning visual cues, multi-scale style learning, and audio-visual fusion. Notably, our model, though not explicitly optimized for this task, shows an emergent capability for generating speech well-aligned with lip movements in given videos.

\begin{figure}[t]
  \centering
  \includegraphics[width=1.0\textwidth]{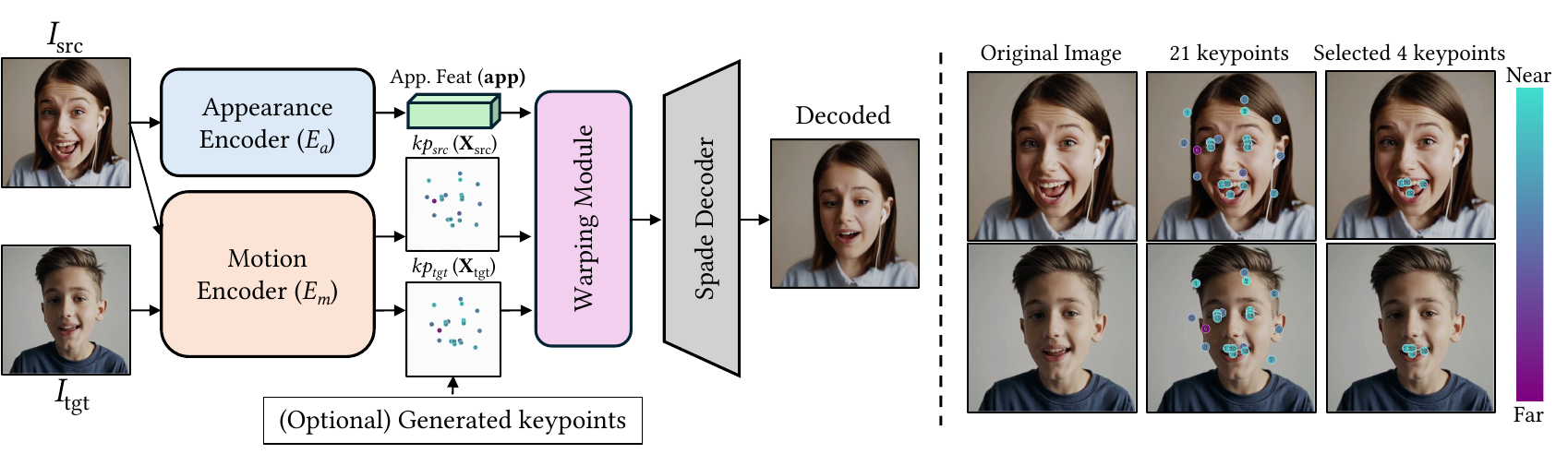}
  \caption{LivePortrait framework and mouth-related expression keypoint analysis. LivePortrait's motion encoder ($E_m$) infers parameters including a 3D expression deformation $\mathbf{e} \in \mathbb{R}^{21 \times 3}$ for 21 canonical keypoints. We find that deforming approximately four specific keypoints (highlighted) primarily dictates mouth articulation. Our Motion-DiT leverages this by generating only the deformation components ($\mathbf{e}^{\text{mouth}} \subset \mathbf{e}$) for these crucial mouth keypoints, enabling efficient lip-sync.}
  \label{fig:liveportrait_vis}
\end{figure}

\section{Preliminaries}
\subsection{LivePortrait Framework}
\label{sec:preliminaries}
LivePortrait~\cite{guo2024liveportrait} enables image-to-video synthesis by disentangling structural and appearance features from a single input image. It employs an appearance encoder $E_a$ to extract a global appearance feature $\mathbf{app} \in \mathbb{R}^{C \times D' \times H' \times W'}$. In parallel, a motion encoder $E_m$ infers a set (21) of 3D implicit facial keypoints $\mathbf{X} \in \mathbb{R}^{21 \times 3}$ which is parametrized by canonical keypoints $\mathbf{x}_c \in \mathbb{R}^{21 \times 3}$, pose matrix $\mathbf{R} \in \mathbb{R}^{3 \times 3}$, expression deformation $\mathbf{e} \in \mathbb{R}^{21 \times 3}$, scale $\mathbf{s}\in \mathbb{R}$ and translation $\mathbf{t} \in \mathbb{R}^{21 \times 3}$. The final keypoints are then computed as: $\mathbf{X} = \mathbf{s} \cdot (\mathbf{x}_{c}\mathbf{R}+\mathbf{e}) + \mathbf{t}$. The warping module modifies $\mathbf{app}$ with estimated keypoint differences and the decoder projects this warped appearance feature $\mathbf{app'}$ into a target video frame.

We analyze this implicit embedding to identify its functional structure. Visualizing the 21 dimensions of the expression embedding $\mathbf{e}$ (\fref{fig:liveportrait_vis}) reveals that approximately four specific dimensions consistently control the mouth region. Isolating these mouth-related components $\mathbf{e}^{\text{mouth}} \subset \mathbf{e}$ and freezing others modifies only the lip shape. This empirical finding suggests lip-sync generation can be simplified by modeling only this small subset of expression dimensions. We leverage this by training our motion generation module to predict only these mouth-related components, which are then combined with fixed identity- and pose-related features for rendering. Importantly, this implicit representation exposes a low-dimensional subspace corresponding to facial expressions and lip shapes within an otherwise high-dimensional facial motion space, which plays a crucial role in enabling near real-time sampling.

\subsection{F5-TTS and Conditional Flow Matching}
F5-TTS~\cite{chen2024f5} is a Conditional Flow Matching (CFM) based text-to-speech model. Inspired by inpainting-based approaches such as VoiceBox~\cite{le2023voicebox}, F5-TTS treats speech generation as a mask-and-predict problem, where parts of the mel-spectrogram are randomly masked during training and then reconstructed conditioned on surrounding context and reference signals. 

Formally, the model receives a masked audio segment $\tilde{\mathbf{a}}^{\text{masked}}$ and a conditioning vector $\mathbf{c}^{\text{text}}$ consisting of unmasked regions, reference audio features, and text embeddings. These are concatenated and passed through a flow-based network trained using the CFM objective described in Eq. \ref{eq:l_cfm}.

This inpainting formulation has two key benefits. First, it enables robust training across diverse conditioning scenarios by randomly varying the masked regions. Second, it inherently models the relationship between unmasked context and the regions to be reconstructed, ensuring consistent speech generation. As a result, F5-TTS produces high-quality, voice-consistent speech for arbitrary text inputs and provides a strong foundation for our Audio-DiT module.


\section{Method}

\subsection{Overview}
Our goal is to simultaneously generate temporally aligned speech audio and facial motion from multimodal inputs (text, reference audio, or motion). To this end, we propose a dual-stream diffusion architecture composed of Audio-DiT and Motion-DiT, partially fused via joint attention blocks. Figure~\ref{fig:architecture} illustrates the overall architecture with training and inference pipeline. 

\subsection{Motion-DiT and Audio-DiT Design and Flow Matching}
\label{sec:motion_dit}

\begin{figure}[t]
  \centering
  \includegraphics[width=1.0\textwidth]{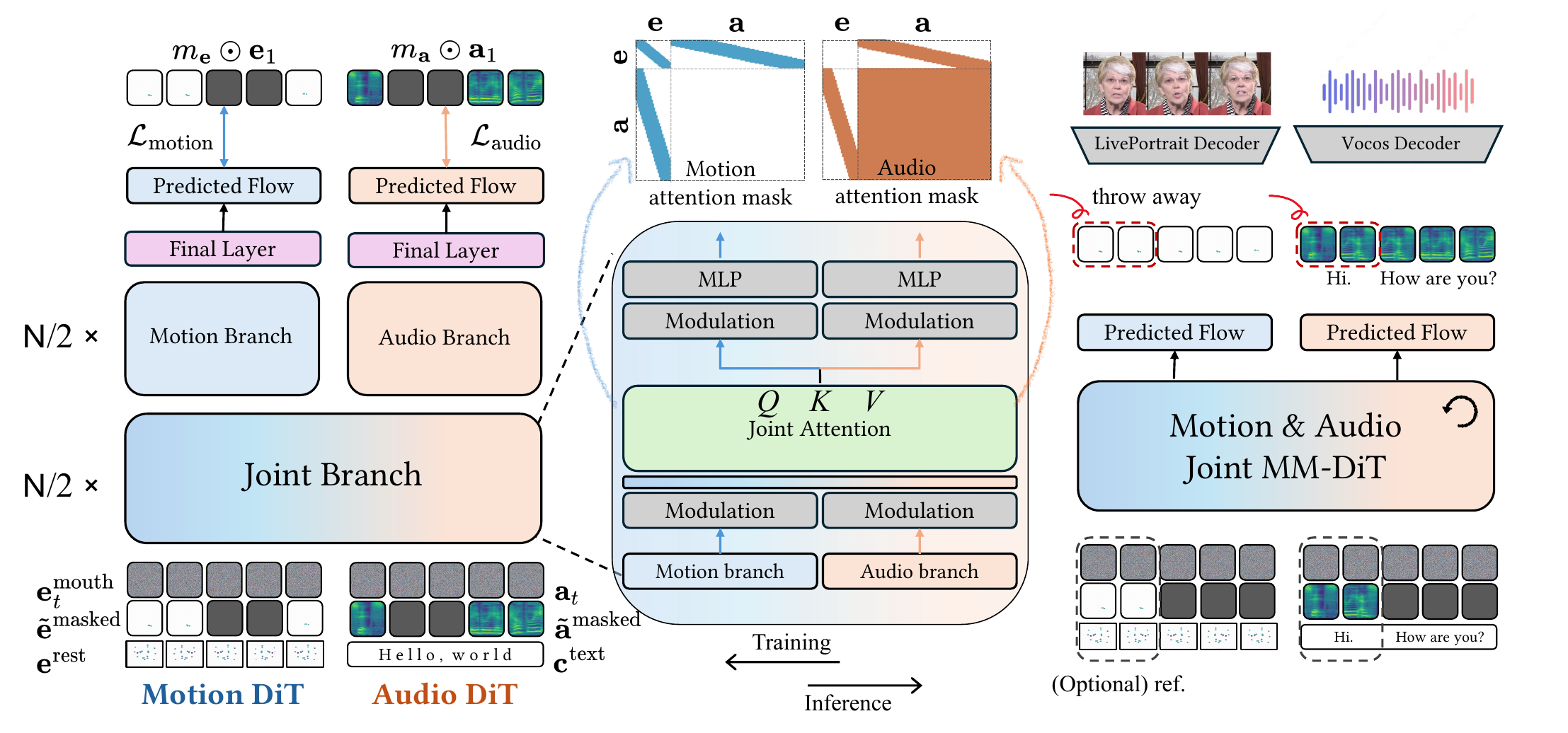}
  \caption{The training and inference pipeline of the JAM-Flow framework. Our joint MM-DiT comprises a Motion-DiT for facial expression keypoints ($\mathbf{e}^{\text{mouth}}$) and an Audio-DiT for mel-spectrograms ($\mathbf{a}$), coupled via joint attention. The model is trained with an inpainting-style flow matching objective on masked inputs and various conditions (text, reference audio/motion). At inference, it flexibly generates synchronized audio-visual outputs from partial inputs.}
  \label{fig:architecture}
\end{figure}

The Motion-DiT generates expression embeddings $\mathbf{e}^{\text{mouth}} \in \mathbb{R}^{T_{\text{frame}} \times 4 \times 3}$ that control lip motion, following the observation from Section~\ref{sec:preliminaries} that 4 of the 21 expression dimensions are sufficient to model mouth dynamics. As shown in Figure~\ref{fig:architecture}, we provide two inputs to the motion stream: the target expression excluding mouth-related components, denoted $\mathbf{e}^{\text{rest}} \in \mathbb{R}^{T_{\text{frame}} \times 17 \times 3 }$ and the audio-derived conditioning features $\mathbf{f}_{\text{audio}}$ from the Audio-DiT stream (if available). During training, the Motion-DiT denoises corrupted $\mathbf{e}^{\text{mouth}}$ vectors via a conditional flow-matching process:
\begin{equation}
    \mathcal{L}_{\text{motion}} = \mathbb{E}_{\mathbf{e}^{\text{mouth}}_0, \mathbf{e}^{\text{mouth}}_1, t} \left[ \| v_\theta(\mathbf{e}^{\text{mouth}}_t, t;\mathbf{f}^{\text{audio}},\tilde{\mathbf{e}}^{\text{masked}},\mathbf{e}^{\text{rest}}) - (\mathbf{e}^{\text{mouth}}_1 - \mathbf{e}^{\text{mouth}}_0) \|^2 \right]
\end{equation}
where $\mathbf{e}_t$ denotes the noisy input at step $t$, and $v_\theta$ is the velocity predicting DiT. This design naturally enables joint generation of two tightly related modalities, while representing lip motion in a low-dimensional space, which is key to efficient and near real-time synthesis.

The Audio-DiT generates mel-spectrograms $\mathbf{a}_{\text{mel}} \in \mathbb{R}^{T_{\text{mel}} \times d_{\text{mel}}}$ using the Conditional Flow Matching (CFM) objective. Following F5-TTS, we apply random inpainting masks to audio segments and condition on reference audio and text features.
Additionally, we adapt the motion-derived conditioning features $\mathbf{f}^{\text{motion}}$ from the Motion-DiT stream (if available).
As defined in Eq.~\ref{eq:l_cfm}, the model learns to reconstruct missing regions from context:
\begin{equation}
    \mathcal{L}_{\text{audio}} = \mathbb{E}_{t, \mathbf{a}_0, \mathbf{a}_1} \left[\|v_\theta(\mathbf{a}_t, t; \mathbf{f}^{\text{motion}}, \tilde{\mathbf{a}}^{\text{masked}}, \mathbf{c}^{\text{text}}) - (\mathbf{a}_1 - \mathbf{a}_0)\|^2\right]
\end{equation}
This enables the Audio-DiT to flexibly operate under partial or complete conditions and naturally generalize to reference audio-based speech generation.

The final flow-matching loss is defined as
\begin{equation}
\mathcal{L} = \mathcal{L}_{\text{audio}} + \mathcal{L}_{\text{motion}}.
\end{equation}
During training, audio and motion streams are randomly masked: sometimes only motion is hidden, sometimes only audio, and at other times both are missing. This stochastic masking compels the model to attend jointly to the available context across modalities, naturally learning dependencies between unmasked and masked regions.

\subsection{Joint Attention and Temporal Fusion}
\label{sec:joint_attention}

To enable cross-modal interactions, we introduce $N_{\text{joint}}$ layers of joint attention between the audio and motion streams, as illustrated in Figure~\ref{fig:architecture}. Tokens from both streams are fused via joint-attention blocks, while the rest of the transformer layers remain modality-specific.

To ensure temporal compatibility, we apply a simple alignment of rotary positional embeddings (RoPE). For position index $p \in [0, L-1]$ and frequency $\theta_d$, the rotation angle is defined as
$\phi(p,d) \;=\; \frac{p}{L}\,L_{\text{ref}} \cdot \theta_d,$
where $L$ is the sequence length of the current modality and $L_{\text{ref}} = \max(L_a, L_m)$. This scaling ensures that audio and motion tokens at the same timestep share comparable angular positions, despite having different sequence lengths. While this adjustment is straightforward, we found it to be indispensable: without such alignment, joint attention fails to converge, as queries and keys from different modalities interact at incompatible positional scales.


\subsection{Attention Masking Strategy}
\label{sec:attn_mask}

A key novelty of our framework lies in the attention masking design for joint modeling. Unlike prior multimodal transformers (e.g., MM-DiT) that apply symmetric attention across streams, we introduce modality-specific asymmetric masks tailored to the functional roles of audio and motion.

For motion tokens, we impose a local self-attention window, reflecting the inductive bias that facial motion depends mainly on temporally adjacent frames. Motion-to-audio attention is restricted to the corresponding local window, while audio-to-audio attention is disabled during motion generation. Conversely, for audio tokens, we retain full global self-attention across the utterance, but restrict cross-modal attention to only motion tokens at the same timestamp, with motion-to-motion self-attention entirely masked out.  (Figure~\ref{fig:architecture}, top)

This design is simple yet crucial: each modality attends in the way most natural to its generative process, while their cross-modal interactions remain tightly aligned. To our knowledge, no prior work has explored such asymmetric, modality-specific masking for joint attention. We found this strategy indispensable for stable training and synchronized outputs.

\subsection{Training Procedure}
Training is conducted in two stages:

\textbf{Stage 1:}
We prepare pretrained models for each modality. For audio, we adopt F5-TTS as the pretrained TTS backbone. For motion, we train a model from scratch following a common practice in talking-head generation, where wav2vec2 features are concatenated to the motion input.

\textbf{Stage 2:}
We then connect the two modalities with joint attention layers, applying RoPE alignment to ensure compatible temporal embeddings. Both branches are jointly trained with gradients flowing across modalities through the shared attention layers, enabling mutual refinement of audio and motion representations. At this stage, wav2vec2 features are no longer used: text conditioning is injected via the Audio-DiT, while the Motion-DiT consumes both $\mathbf{e}_{\text{rest}}$ and intermediate audio features $\mathbf{f}^{\text{audio}}$. Further architectural and training details are provided in the supplementary material.

\section{Experiments}
\subsection{Experimental Setup}
We train our model on the CelebV-Dub~\cite{sung2025voicecraft} dataset, filtered from CelebV-HQ~\cite{zhu2022celebv} and CelebV-Text~\cite{yu2023celebv}. Training is conducted in two stages: the Motion-DiT is first trained from scratch using keypoint representations, while the Audio-DiT is initialized from F5-TTS. After convergence, joint training is performed on paired audio-visual data. Evaluation is carried out on CelebV-Dub test splits and HDTF~\cite{zhang2021flow}, using standard metrics including WER, SIM-o, LSE-C, LSE-D, spkSIM, FID, and FVD. During our experiments, we found that CelebV-Dub is annotated with \textit{Whisper-generated pseudo-transcripts} rather than ground-truth transcripts, introducing nontrivial label noise; as a result, performance can be slightly lower than that of dedicated TTS models trained on clean GT pairs.

\subsection{Quantitative Evaluation}
\begin{table*}[t]
\centering
\caption{Talking-head generation comparison on HDTF~\cite{zhang2021flow}. \textbf{Inference} is measured on a \textbf{single RTX A6000} for generating a \textbf{20-second} video, unless otherwise noted. $^{\dagger}$SadTalker is reported with GFPGAN. $^{\ddagger}$Hallo3 is reported on an \textbf{H100} GPU.  With additional engineering and an H100-class GPU, our method can approach real-time generation (19s for a 20s video at \textbf{25 fps}).}
\label{tab:talkinghead}
\footnotesize
\setlength{\tabcolsep}{3pt}
\resizebox{\textwidth}{!}{%
\begin{tabular}{lcccccc}
\toprule
\textbf{Method} & \textbf{FID ↓} & \textbf{FVD ↓} & \textbf{LSE-C ↑} & \textbf{LSE-D ↓} & \textbf{Inference (20s) ↓} & \textbf{\#Params} \\
\midrule
Ground Truth & - & - & 8.70 & 6.597 & - & - \\
\midrule
SadTalker~\cite{zhang2023sadtalker}       & 22.340     & 203.860     & 7.885     & 7.545 & 2.5 min$^{\dagger}$ & $\sim$200M \\
DreamTalk~\cite{ma2023dreamtalk}          & 78.147     & 890.660     & 6.376     & 8.364 & - & - \\
AniPortrait~\cite{wei2024aniportrait}     & 26.561     & 234.666     & 4.015     & 10.548 & 7 min & $\sim$1B \\
Hallo~\cite{xu2024hallo}                  & 20.545     & 173.497     & 7.750     & 7.659 & 23 min & 1--3B \\
Hallo3~\cite{cui2025hallo3}               & 20.359     & 160.838     & 7.252     & 8.106 & 30 min$^{\ddagger}$ & 1--3B \\
Ours-Stage1                               & 18.372     & 194.27      & 7.138     & 7.947 & - & - \\
\midrule
Ours (I2V)                                & 17.571     & 192.30      & 7.324     & 7.777 & 45 sec & $\sim$500M (joint) \\
Ours (V2V)                                & \textbf{11.633} & \textbf{25.07} & \textbf{8.086} & \textbf{7.181} & 45 sec & $\sim$500M (joint) \\
\bottomrule
\end{tabular}
}

\end{table*}

Our model is the first to be explicitly trained for joint audio–motion generation, and thus there are no direct baselines for this unified setting. Nevertheless, by controlling which inputs remain unmasked at inference, the same model can flexibly perform multiple tasks, for example, audio-only, motion-only, or full audiovisual generation. To demonstrate its versatility, we evaluate across three representative tasks by comparing with models that are individually specialized for each, emphasizing that our approach does not rely on task-specific architectures yet still achieves competitive results.
Because common objective metrics may not fully reflect perceptual quality and synchronization---especially under codec variations or TTS-generated audio---we additionally report a user study to better capture human preferences (Sec.~\ref{appedix:user}).

\paragraph{Talking head generation.}
\begin{wraptable}{R}{0.45\textwidth}
\vspace{-1em}
\centering
\caption{Comparison of text-to-speech performance in LibriSpeech-PC test-clean~\cite{chen2024f5, panayotov2015librispeech} benchmark. Methods are from prior work~\cite{du2024cosyvoice, guo2024fireredtts, eskimez2024e2, chen2024f5, jiang2025megatts}}
\label{tab:tts}
\footnotesize
\setlength{\tabcolsep}{4pt}
\begin{tabular}{lcc}
    \toprule
    \textbf{Method} & \textbf{WER ↓} & \textbf{SIM-o ↑} \\
    \midrule
    Cosy Voice      & 3.59\%      & 0.66      \\
    FireRedTTS      & 2.69\%      & 0.47      \\
    E2 TTS          & 2.95\%      & 0.69      \\
    F5-TTS          & 2.42\%      & 0.66      \\
    MegaTTS 3       & \textbf{2.31\%}      & \textbf{0.70}      \\
    \midrule
    Ours            & 4.91\% & 0.64 \\
    \bottomrule
\end{tabular}
\vspace{-1em}
\end{wraptable}
We evaluate talking head performance on HDTF using four key metrics: FID, FVD, LSE-C, and LSE-D. As shown in Table~\ref{tab:talkinghead}, our model performs competitively or better with SOTA methods such as SadTalker, AniPortrait, and Hallo3. Notably, our method is primarily designed with a video-to-video (V2V) setup in mind, utilizing a sequence of 17 non-mouth expression keypoints as a `clue' (following~\cite{zhong2023identity}). This keypoint `clue' is used in both V2V and image-to-video (I2V) configurations: V2V warps frames sequentially from a source video, whereas I2V consistently warps an initial source image.

In addition to quality, our approach is substantially faster: on a single RTX A6000, it generates a 20-second video in 45 seconds, and with additional engineering on an H100-class GPU it can approach real-time throughput (e.g., 19 s for 20 s at 25 fps; see Table~\ref{tab:talkinghead}).
Moreover, this quantitative improvement aligns with human perception: in our user study on HDTF, participants ranked our V2V and I2V variants as the top two methods in overall quality (Fig.~\ref{fig:userstudy-th}).

\paragraph{Text-to-speech generation.}
TTS performance is evaluated using WER and SIM-o on LibriSpeech-PC test-clean~\cite{chen2024f5, panayotov2015librispeech}. Although our model is trained for joint audio--motion generation, for a fair comparison against dedicated TTS systems we perform TTS evaluation by generating \emph{audio only} at inference time. As shown in Table~\ref{tab:tts}, our WER is slightly higher than that of dedicated TTS systems. We stress, however, that our model is not optimized as a pure TTS system; it is a unified audio--motion generator trained under a substantially noisier supervision regime, where WER is largely \emph{data-bounded} rather than \emph{capacity-bounded}.

Concretely, CelebV-Dub provides no human-verified transcripts, and we therefore rely on Whisper-generated pseudo-GT captions. Transcription errors in Whisper directly propagate to the supervision signal and impose a lower bound on attainable WER. This is consistent with prior findings on LibriSpeech-PC: the reported WER of Whisper-base is 4.50\%~\cite{chen2024f5}, which closely matches the $\sim$4.9\% WER we observe, indicating that label noise is the primary bottleneck. Moreover, the audio track in CelebV-Dub is obtained via source separation (Spleeter), which introduces artifacts absent in standard audio-only TTS corpora and further limits achievable fidelity.

Finally, we caution that WER is not a reliable proxy for perceptual quality, especially in our multimodal setting: small transcript-level errors may have limited impact on intelligibility or speaker similarity, and recent work has highlighted the mismatch between WER and human judgments of transcript readability and severity of errors~\cite{saon17_interspeech}. Taken together, these factors explain the modest WER gap to specialized TTS baselines, while our qualitative results demonstrate that joint training captures cross-modal relationships without a meaningful degradation in perceived audio quality.

\paragraph{Automated video dubbing.}

\begin{table}[t]
\centering
\caption{Comparison of automated video dubbing performance in CelebV-Dub~\cite{sung2025voicecraft} dataset. Methods are from prior work~\cite{peng2024voicecraft, cong2023learning, cong2024styledubber, sung2025voicecraft}}
\label{tab:dubbing}
\footnotesize
\setlength{\tabcolsep}{4pt}
\resizebox{0.75\textwidth}{!}{%
\begin{tabular}{@{}lcccc@{}}
    \toprule
    \textbf{Method} & \textbf{LSE-C ↑} & \textbf{LSE-D ↓} & \textbf{WER ↓} & \textbf{spkSIM ↑} \\
    \midrule
    Ground Truth    & 6.73             & 7.44             & 4.15\%         & -                \\
    Zero-Shot TTS   & 2.78             & 11.68            & 3.83\%         & 0.316            \\
    \midrule
    HPMDubbing      & \textbf{6.36}    & \textbf{7.80}    & 24.06\%        & 0.146            \\
    StyleDubber     & 3.78             & 10.40            & 9.48\%         & 0.264            \\
    VoiceCraft-Dub  & 6.05             & 8.33             & 7.01\%         & 0.333            \\
    \midrule
    Ours            & 3.43             & 10.56            & \textbf{6.39\%}& \textbf{0.410}   \\
    \bottomrule
\end{tabular}
}
\end{table}
Thanks to the inpainting-based training paradigm, our model naturally extends to automated video dubbing, generating speech that is both semantically correct and temporally aligned with the speaker’s lip movements. However, as discussed by~\cite{yaman2024audio, muaz2023sidgan, sung2025voicecraft}, we found that SyncNet-derived metrics~\cite{chung2017out} (LSE-C, LSE-D) often fail to operate reliably, showing instability under codec variations and, most critically, with TTS outputs from our model. We therefore caution against over-interpreting these scores and refer readers to the supplementary materials for qualitative examples that better reflect performance. As seen in Table~\ref{tab:dubbing}, our model achieves strong WER and the highest spkSIM among all methods, demonstrating its effectiveness for this task.
To complement these imperfect objective metrics, we also conduct a user study on CelebV-Dub, where our method is preferred in 62.6\% of cases over VoiceCraft-Dub (37.4\%), while other baselines receive no votes (Fig.~\ref{fig:userstudy-dubbing}).

\subsection{Qualitative Results}
We provide a supplementary webpage with videos. On standard settings (talking‐head comparison, TTS comparison, and automated dubbing), our method shows stronger lip–audio alignment and speaker consistency; please see the supplementary videos for side-by-side examples and details.

\paragraph{Our Exclusive use cases.}
\begin{enumerate}
\item Text $\rightarrow$ Audio + Motion: Without any reference audio, the model \emph{jointly} generates speech with randomly sampled speaker identity and facial motion that remain tightly synchronized. This text-only joint generation is a primary target of our training.

\item Text + Reference Audio $\rightarrow$ Audio + Motion: Given target text and a short voice reference, the model co-generates speech in the reference voice and the matching lip motion. This is likewise a core \emph{joint} audio–motion scenario we explicitly train for.

\item Reference Motion + Target Text $\rightarrow$ Audio: With the video fixed, the synthesized audio adapts its opening/closure timing to the observed lip motion, even when perfect articulation is impossible, indicating active attention to motion during speech generation.

\item Reference Motion $\rightarrow$ Audio (no text): Supplying only motion still yields plausible, time-aligned speech, showing an implicit mapping from visual articulation to acoustics under partial conditioning.

\end{enumerate}
These results highlight the controllability and robustness of our unified model under partially missing inputs; see the supplementary webpage for qualitative examples.




\begin{table}[t]
\centering
\caption{Ablation study on the number of joint attention blocks, motion attention masking, and Audio-DiT finetuning.}
\label{tab:ablation}
\scriptsize
\setlength{\tabcolsep}{2.5pt}

\begin{tabular*}{\textwidth}{@{\extracolsep{\fill}}ccc ccccc@{}}

\toprule
\makecell{\textbf{\# Joint}\\\textbf{Blocks}} &
\makecell{\textbf{Attn.}\\\textbf{Mask}} &
\makecell{\textbf{Train}\\\textbf{Audio-DiT}} &
\textbf{FID ↓} &
\textbf{LSE-C ↑} &
\textbf{LSE-D ↓} &
\textbf{WER ↓} &
\textbf{SIM-o ↑} \\
\midrule

22 (Full)    & \xmark & \xmark & 5.759 & 4.81 & 8.40 & 6.88\% & 0.62 \\
11 (Half)    & \xmark & \xmark & 5.735 & 4.64 & 9.07 & 7.25\% & 0.62 \\
11 (Half)    & \xmark & \cmark & 5.748 & 6.44 & 7.99 & 7.93\% & 0.61 \\
11 (Half)    & \cmark & \xmark & 5.747 & 5.76 & 8.22 & \textbf{6.76\%} & 0.63 \\
11 (Half)    & \cmark & \cmark & \textbf{5.662} & \textbf{6.45} & \textbf{7.73} & 7.28\% & \textbf{0.64} \\

\bottomrule
\end{tabular*}
\end{table}

\subsection{Ablation Studies}
To assess the impact of our design choices, we conduct controlled ablation experiments along three axes: (1) the degree of joint attention between audio and motion streams, (2) the presence of temporal attention masks in the Motion-DiT, and (3) the finetuning of the Audio-DiT during stage-2 training. Given the aforementioned instability of LSE-C and LSE-D with generated audio, and to better assess the commonly adopted audio-driven talking head setup, we calculate LSE-C and LSE-D in the ablation study using generated motion paired with ground truth (GT) audio. WER and SIM-o are computed using both generated motion and audio, while LSE-C and LSE-D use GT audio for stability.

\paragraph{Joint Attention Configuration.}
We compared a \textit{Full Joint} configuration (all DiT layers share attention) with our \textit{Half Joint} approach (only earlier layers fused). Although Table~\ref{tab:ablation} shows numerically stronger scores for Full Joint, it was both unstable and computationally much heavier. In practice, we found that Half Joint provided a more reliable trade-off: comparable qualitative performance with significantly lower training cost. Our choice of Half Joint therefore reflects a balance between performance and efficiency, which proved most practical for large-scale experiments.

\paragraph{Attention Masking (and RoPE).}
Removing temporal attention masks (\textit{No Masking}) leads to sharp drops in lip-sync quality. While the degradation may not always be dramatic in raw scores, subjectively the model often fails to achieve coherent joint training at all, producing temporally drifting or unsynchronized motion. In other words, masking is not just helpful but almost indispensable for stable learning. Similarly, RoPE alignment (Section~\ref{sec:joint_attention}) is absolutely critical: without it, joint training does not converge at all, which is why we do not report a corresponding ablation in the table.

\paragraph{Audio-DiT Finetuning.}
We also tested freezing the Audio-DiT during stage-2 training to preserve the strong F5-TTS prior. As shown in Table~\ref{tab:tts} and Table~\ref{tab:ablation}, this setting slightly improves WER but results in weaker synchronization. Our experiments revealed why: keeping Audio-DiT fixed prevents the system from learning a proper joint audio–motion distribution, leaving Motion-DiT to adapt alone. Allowing Audio-DiT to finetune, by contrast, enables both modalities to co-adapt through the shared attention layers, yielding more synchronized and coherent outputs.

Taken together, our experiments show that Half Joint attention, modality-specific masking with RoPE alignment, and Audio-DiT finetuning are essential not only for quantitative gains but also for stable and efficient joint training that truly learns a shared multimodal distribution.

\section{Discussion and Ethics}

Our experiments show that partial joint attention and localized temporal masking are key to stable and coherent multimodal generation. While full joint attention yields slightly better scores, it often results in unstable training. Our hybrid design balances cross-modal fusion and modality-specific representations, contributing to both robustness and performance. 
We further find that the compact LivePortrait keypoint warping is not only computationally efficient but also preserves identity and fine-grained facial details; in contrast, diffusion-based talking-head methods (e.g., Hallo/Hallo3) often exhibit noticeable temporal flicker in longer sequences despite handling complex scenes well.

We also observed that the generated speech often reflects emotional cues from facial motion. For instance, smiling motions tend to produce brighter, higher-pitched voices, despite the absence of explicit emotion supervision. This suggests the model implicitly aligns emotion across modalities, opening possibilities for expressive and emotionally-aware generation. More broadly, the generated audio exhibits coherent prosody with facial motion (e.g., natural pauses during speech breaks and consistent tone shifts), suggesting that joint training captures cross-modal alignment beyond simple temporal synchronization.

A major strength of our model is its versatility. It supports diverse input configurations (e.g., audio-only, motion-only, text + portrait) within a single framework, unlike prior models limited to either talking head synthesis or TTS. This flexibility enables applications such as expressive dubbing, silent video revoicing, and adaptive avatar generation. Interestingly, during simultaneous audio and motion generation, we observe an asymmetric adaptation: the model tends to preserve visual consistency with minimal changes to lip motion, while adjusting the audio more substantially to match the given motion and achieve accurate lip-sync. This behavior likely reflects the relative difficulty of modifying realistic visual motion compared to audio synthesis, and provides a practical mechanism for maintaining visual identity while improving synchronization.

Beyond our current scope, we found that generating two modalities jointly is remarkably effective, and we believe this paradigm can naturally extend to other modality pairs (e.g., depth + video, audio + video). While recent concurrent works have started to explore related directions in native multimodal generation, particularly in the video-audio domain~\cite{hacohen2026ltx, veo3_2026, wang2026klear}, our findings further highlight the effectiveness of inpainting-style joint supervision as a simple and scalable mechanism for learning cross-modal relationships, as evidenced in our joint audio-motion setting. More broadly, we believe this paradigm merits deeper exploration as a general framework for multimodal co-generation.

The JAM-Flow model also raises ethical considerations. While it provides benefits for accessibility, avatars, and creative tools, it may also be misused for deepfakes or amplify biases in the data. To address these risks, we plan to explore safeguards such as watermarking and continue reflecting on responsible deployment as the technology evolves.

\section{Conclusion}

We presented the first joint training framework for speech and facial motion generation, integrating model architecture and training strategy to enable mutual conditioning across modalities. Our unified flow-matching-based design combines modality-specific DiT modules with selectively applied joint attention, RoPE alignment, and attention masking, producing coherent multimodal synthesis without separate pipelines.
In addition, by analyzing and leveraging pretrained representation embeddings, we design the system to enable efficient, near real-time sampling.

Through extensive experiments, we demonstrated that inpainting-style joint supervision effectively learns cross-modal relations and supports flexible inference scenarios such as motion-to-audio, audio-to-motion, and full generation from text and portrait image alone, suggesting it as a powerful paradigm for multimodal co-generation.
Finally, we provided a practical architectural pathway for connecting pretrained unimodal models, showing that partial joint attention, RoPE alignment, and localized masking together balance stability, performance, and efficiency, yielding competitive results with specialized state-of-the-art models.

While these findings are promising, current limitations stem largely from data and compute. CelebV-Dub~\cite{sung2025voicecraft}, for instance, contains Whisper-generated captions with transcription errors and demuxed audio with artifacts, while LivePortrait constrains motion modeling largely to facial regions. Nevertheless, we believe our results make a compelling case for unified multimodal training. With more curated datasets and stronger video diffusion backbones~\cite{yang2024cogvideox, hacohen2024ltx, wan2025wan}, future work can further extend this framework, unlocking expressive dubbing, controllable avatars, and more natural human–computer interaction.


\clearpage
%
%
\bibliographystyle{splncs04}
\bibliography{main}

@STRING{cvpr = {IEEE Conference on Computer Vision and Pattern Recognition (CVPR)}}

@STRING{iccv = {IEEE International Conference on Computer Vision (ICCV)}}

@STRING{eccv = {European Conference on Computer Vision (ECCV)}}

@STRING{cvprw = {IEEE Conference on Computer Vision and Pattern Recognition (CVPR) Workshop}}

@STRING{iclr = {International Conference on Learning Representations (ICLR)}}

@STRING{neurips = {Conference on Neural Information Processing Systems (NeurIPS)}}

@STRING{icml = {International Conference on Machine Learning (ICML)}}

@STRING{sig = {ACM SIGGRAPH}}

@STRING{siga = {ACM SIGGRAPH Asia}}

@article{wan2025wan,
  title={Wan: Open and advanced large-scale video generative models},
  author={Wan, Team and Wang, Ang and Ai, Baole and Wen, Bin and Mao, Chaojie and Xie, Chen-Wei and Chen, Di and Yu, Feiwu and Zhao, Haiming and Yang, Jianxiao and others},
  journal={arXiv preprint arXiv:2503.20314},
  year={2025}
}

@article{hacohen2024ltx,
  title={Ltx-video: Realtime video latent diffusion},
  author={HaCohen, Yoav and Chiprut, Nisan and Brazowski, Benny and Shalem, Daniel and Moshe, Dudu and Richardson, Eitan and Levin, Eran and Shiran, Guy and Zabari, Nir and Gordon, Ori and others},
  journal={arXiv preprint arXiv:2501.00103},
  year={2024}
}

@inproceedings{panayotov2015librispeech,
  title={Librispeech: an asr corpus based on public domain audio books},
  author={Panayotov, Vassil and Chen, Guoguo and Povey, Daniel and Khudanpur, Sanjeev},
  booktitle={2015 IEEE international conference on acoustics, speech and signal processing (ICASSP)},
  pages={5206--5210},
  year={2015},
  organization={IEEE}
}

@inproceedings{chung2017out,
  title={Out of time: automated lip sync in the wild},
  author={Chung, Joon Son and Zisserman, Andrew},
  booktitle={Computer Vision--ACCV 2016 Workshops: ACCV 2016 International Workshops, Taipei, Taiwan, November 20-24, 2016, Revised Selected Papers, Part II 13},
  pages={251--263},
  year={2017},
  organization={Springer}
}

@inproceedings{muaz2023sidgan,
  title={Sidgan: High-resolution dubbed video generation via shift-invariant learning},
  author={Muaz, Urwa and Jang, Wondong and Tripathi, Rohun and Mani, Santhosh and Ouyang, Wenbin and Gadde, Ravi Teja and Gecer, Baris and Elizondo, Sergio and Madad, Reza and Nair, Naveen},
  booktitle=iccv,
  year={2023}
}

@inproceedings{yaman2024audio,
  title={Audio-Visual Speech Representation Expert for Enhanced Talking Face Video Generation and Evaluation},
  author={Yaman, Dogucan and Eyiokur, Fevziye Irem and B{\"a}rmann, Leonard and Akti, Seymanur and Ekenel, Haz{\i}m Kemal and Waibel, Alexander},
  booktitle=cvprw,
  year={2024}
}

@inproceedings{zhang2021flow,
  title={Flow-guided one-shot talking face generation with a high-resolution audio-visual dataset},
  author={Zhang, Zhimeng and Li, Lincheng and Ding, Yu and Fan, Changjie},
  booktitle=cvpr,
  year={2021}
}

@inproceedings{zhu2022celebv,
  title={CelebV-HQ: A large-scale video facial attributes dataset},
  author={Zhu, Hao and Wu, Wayne and Zhu, Wentao and Jiang, Liming and Tang, Siwei and Zhang, Li and Liu, Ziwei and Loy, Chen Change},
  booktitle=eccv,
  year={2022},
}

@article{ma2023dreamtalk,
  title={Dreamtalk: When expressive talking head generation meets diffusion probabilistic models},
  author={Ma, Yifeng and Zhang, Shiwei and Wang, Jiayu and Wang, Xiang and Zhang, Yingya and Deng, Zhidong},
  journal={arXiv preprint arXiv:2312.09767},
  volume={2},
  number={3},
  year={2023}
}

@article{cong2024styledubber,
  title={Styledubber: towards multi-scale style learning for movie dubbing},
  author={Cong, Gaoxiang and Qi, Yuankai and Li, Liang and Beheshti, Amin and Zhang, Zhedong and Hengel, Anton van den and Yang, Ming-Hsuan and Yan, Chenggang and Huang, Qingming},
  journal={arXiv preprint arXiv:2402.12636},
  year={2024}
}

@inproceedings{cong2023learning,
  title={Learning to dub movies via hierarchical prosody models},
  author={Cong, Gaoxiang and Li, Liang and Qi, Yuankai and Zha, Zheng-Jun and Wu, Qi and Wang, Wenyu and Jiang, Bin and Yang, Ming-Hsuan and Huang, Qingming},
  booktitle=cvpr,
  year={2023}
}

@article{peng2024voicecraft,
  title={Voicecraft: Zero-shot speech editing and text-to-speech in the wild},
  author={Peng, Puyuan and Huang, Po-Yao and Li, Shang-Wen and Mohamed, Abdelrahman and Harwath, David},
  journal={arXiv preprint arXiv:2403.16973},
  year={2024}
}

@article{du2024cosyvoice,
  title={Cosyvoice: A scalable multilingual zero-shot text-to-speech synthesizer based on supervised semantic tokens},
  author={Du, Zhihao and Chen, Qian and Zhang, Shiliang and Hu, Kai and Lu, Heng and Yang, Yexin and Hu, Hangrui and Zheng, Siqi and Gu, Yue and Ma, Ziyang and others},
  journal={arXiv preprint arXiv:2407.05407},
  year={2024}
}

@article{guo2024fireredtts,
  title={Fireredtts: A foundation text-to-speech framework for industry-level generative speech applications},
  author={Guo, Hao-Han and Hu, Yao and Liu, Kun and Shen, Fei-Yu and Tang, Xu and Wu, Yi-Chen and Xie, Feng-Long and Xie, Kun and Xu, Kai-Tuo},
  journal={arXiv preprint arXiv:2409.03283},
  year={2024}
}

@article{sung2025voicecraft,
  title={{VoiceCraft-Dub: Automated Video Dubbing with Neural Codec Language Models}},
  author={Sung-Bin, Kim and Choi, Jeongsoo and Peng, Puyuan and Chung, Joon Son and Oh, Tae-Hyun and Harwath, David},
  journal={arXiv preprint arXiv:2504.02386},
  year={2025}
}

@article{chen2024f5,
  title={F5-tts: A fairytaler that fakes fluent and faithful speech with flow matching},
  author={Chen, Yushen and Niu, Zhikang and Ma, Ziyang and Deng, Keqi and Wang, Chunhui and Zhao, Jian and Yu, Kai and Chen, Xie},
  journal={arXiv preprint arXiv:2410.06885},
  year={2024}
}

@inproceedings{eskimez2024e2,
  title={E2 tts: Embarrassingly easy fully non-autoregressive zero-shot tts},
  author={Eskimez, Sefik Emre and Wang, Xiaofei and Thakker, Manthan and Li, Canrun and Tsai, Chung-Hsien and Xiao, Zhen and Yang, Hemin and Zhu, Zirun and Tang, Min and Tan, Xu and others},
  booktitle={2024 IEEE Spoken Language Technology Workshop (SLT)},
  pages={682--689},
  year={2024},
  organization={IEEE}
}

@article{lee2024ditto,
  title={Ditto-tts: Efficient and scalable zero-shot text-to-speech with diffusion transformer},
  author={Lee, Keon and Kim, Dong Won and Kim, Jaehyeon and Cho, Jaewoong},
  journal={arXiv preprint arXiv:2406.11427},
  volume={1},
  year={2024}
}

@article{ju2024naturalspeech,
  title={Naturalspeech 3: Zero-shot speech synthesis with factorized codec and diffusion models},
  author={Ju, Zeqian and Wang, Yuancheng and Shen, Kai and Tan, Xu and Xin, Detai and Yang, Dongchao and Liu, Yanqing and Leng, Yichong and Song, Kaitao and Tang, Siliang and others},
  journal={arXiv preprint arXiv:2403.03100},
  year={2024}
}

@article{shen2023naturalspeech,
  title={Naturalspeech 2: Latent diffusion models are natural and zero-shot speech and singing synthesizers},
  author={Shen, Kai and Ju, Zeqian and Tan, Xu and Liu, Yanqing and Leng, Yichong and He, Lei and Qin, Tao and Zhao, Sheng and Bian, Jiang},
  journal={arXiv preprint arXiv:2304.09116},
  year={2023}
}

@article{tan2024naturalspeech,
  title={Naturalspeech: End-to-end text-to-speech synthesis with human-level quality},
  author={Tan, Xu and Chen, Jiawei and Liu, Haohe and Cong, Jian and Zhang, Chen and Liu, Yanqing and Wang, Xi and Leng, Yichong and Yi, Yuanhao and He, Lei and others},
  journal={IEEE Transactions on Pattern Analysis and Machine Intelligence},
  volume={46},
  number={6},
  pages={4234--4245},
  year={2024},
  publisher={IEEE}
}

@article{wang2023neural,
  title={Neural codec language models are zero-shot text to speech synthesizers},
  author={Wang, Chengyi and Chen, Sanyuan and Wu, Yu and Zhang, Ziqiang and Zhou, Long and Liu, Shujie and Chen, Zhuo and Liu, Yanqing and Wang, Huaming and Li, Jinyu and others},
  journal={arXiv preprint arXiv:2301.02111},
  year={2023}
}

@article{ren2020fastspeech,
  title={Fastspeech 2: Fast and high-quality end-to-end text to speech},
  author={Ren, Yi and Hu, Chenxu and Tan, Xu and Qin, Tao and Zhao, Sheng and Zhao, Zhou and Liu, Tie-Yan},
  journal={arXiv preprint arXiv:2006.04558},
  year={2020}
}

@article{ren2019fastspeech,
  title={{Fastspeech: Fast, robust and controllable text to speech}},
  author={Ren, Yi and Ruan, Yangjun and Tan, Xu and Qin, Tao and Zhao, Sheng and Zhao, Zhou and Liu, Tie-Yan},
  journal=neurips,
  year={2019}
}

@article{wang2017tacotron,
  title={Tacotron: Towards end-to-end speech synthesis},
  author={Wang, Yuxuan and Skerry-Ryan, RJ and Stanton, Daisy and Wu, Yonghui and Weiss, Ron J and Jaitly, Navdeep and Yang, Zongheng and Xiao, Ying and Chen, Zhifeng and Bengio, Samy and others},
  journal={arXiv preprint arXiv:1703.10135},
  year={2017}
}

@article{tian2025emo2,
  title={EMO2: End-Effector Guided Audio-Driven Avatar Video Generation},
  author={Tian, Linrui and Hu, Siqi and Wang, Qi and Zhang, Bang and Bo, Liefeng},
  journal={arXiv preprint arXiv:2501.10687},
  year={2025}
}

@article{xu2024hallo,
  title={{Hallo: Hierarchical audio-driven visual synthesis for portrait image animation}},
  author={Xu, Mingwang and Li, Hui and Su, Qingkun and Shang, Hanlin and Zhang, Liwei and Liu, Ce and Wang, Jingdong and Yao, Yao and Zhu, Siyu},
  journal={arXiv preprint arXiv:2406.08801},
  year={2024}
}

@article{yang2024cogvideox,
  title={{Cogvideox: Text-to-video diffusion models with an expert transformer}},
  author={Yang, Zhuoyi and Teng, Jiayan and Zheng, Wendi and Ding, Ming and Huang, Shiyu and Xu, Jiazheng and Yang, Yuanming and Hong, Wenyi and Zhang, Xiaohan and Feng, Guanyu and others},
  journal={arXiv preprint arXiv:2408.06072},
  year={2024}
}

@inproceedings{zhong2023identity,
  title={Identity-preserving talking face generation with landmark and appearance priors},
  author={Zhong, Weizhi and Fang, Chaowei and Cai, Yinqi and Wei, Pengxu and Zhao, Gangming and Lin, Liang and Li, Guanbin},
  booktitle=cvpr,
  year={2023}
}

@inproceedings{tewari2020stylerig,
  title={{Stylerig: Rigging stylegan for 3d control over portrait images}},
  author={Tewari, Ayush and Elgharib, Mohamed and Bharaj, Gaurav and Bernard, Florian and Seidel, Hans-Peter and P{\'e}rez, Patrick and Zollhofer, Michael and Theobalt, Christian},
  booktitle=cvpr,
  pages={6142--6151},
  year={2020}
}

@article{xu2024vasa,
  title={{Vasa-1: Lifelike audio-driven talking faces generated in real time}},
  author={Xu, Sicheng and Chen, Guojun and Guo, Yu-Xiao and Yang, Jiaolong and Li, Chong and Zang, Zhenyu and Zhang, Yizhong and Tong, Xin and Guo, Baining},
  journal=neurips,
  year={2024}
}

@inproceedings{tian2024emo,
  title={{Emo: Emote portrait alive generating expressive portrait videos with audio2video diffusion model under weak conditions}},
  author={Tian, Linrui and Wang, Qi and Zhang, Bang and Bo, Liefeng},
  booktitle=eccv,
  year={2024},
  organization={Springer}
}

@inproceedings{drobyshev2024emoportraits,
  title={{Emoportraits: Emotion-enhanced multimodal one-shot head avatars}},
  author={Drobyshev, Nikita and Casademunt, Antoni Bigata and Vougioukas, Konstantinos and Landgraf, Zoe and Petridis, Stavros and Pantic, Maja},
  booktitle=cvpr,
  year={2024}
}

@inproceedings{xie2024x,
  title={{X-portrait: Expressive portrait animation with hierarchical motion attention}},
  author={Xie, You and Xu, Hongyi and Song, Guoxian and Wang, Chao and Shi, Yichun and Luo, Linjie},
  booktitle=sig,
  year={2024}
}

@article{wei2024aniportrait,
  title={{Aniportrait: Audio-driven synthesis of photorealistic portrait animation}},
  author={Wei, Huawei and Yang, Zejun and Wang, Zhisheng},
  journal={arXiv preprint arXiv:2403.17694},
  year={2024}
}

@article{siarohin2019first,
  title={{First order motion model for image animation}},
  author={Siarohin, Aliaksandr and Lathuili{\`e}re, St{\'e}phane and Tulyakov, Sergey and Ricci, Elisa and Sebe, Nicu},
  journal=neurips,
  year={2019}
}

@article{guo2024liveportrait,
  title={Liveportrait: Efficient portrait animation with stitching and retargeting control},
  author={Guo, Jianzhu and Zhang, Dingyun and Liu, Xiaoqiang and Zhong, Zhizhou and Zhang, Yuan and Wan, Pengfei and Zhang, Di},
  journal={arXiv preprint arXiv:2407.03168},
  year={2024}
}

@inproceedings{prajwal2020wav2lip,
  title={A lip sync expert is all you need for speech to lip generation in the wild},
  author={Prajwal, KR and Mukhopadhyay, Rudrabha and Namboodiri, Vinay P and Jawahar, CV},
  booktitle={Proceedings of the 28th ACM international conference on multimedia},
  pages={484--492},
  year={2020}
}

@inproceedings{zhou2020makeittalk,
  author    = {Yang Zhou and Xintong Han and Eli Shechtman and Jose Echevarria and Evangelos Kalogerakis and Dingzeyu Li},
  title     = {{MakeItTalk}: Speaker-Aware Talking-Head Animation},
  booktitle = siga,
  year      = {2020}
}

@inproceedings{zhang2023sadtalker,
  author    = {Wenxuan Zhang and Xiaodong Cun and Xuan Wang and Yong Zhang and Xi Shen and Yu Guo and Ying Shan and Fei Wang},
  title     = {{SadTalker}: Learning Realistic 3D Motion Coefficients for Stylized Audio-Driven Single Image Talking Face Animation},
  booktitle = cvpr,
  year      = {2023},
}

@article{lin2025omnihuman,
  author    = {Gaojie Lin and Jianwen Jiang and Jiaqi Yang and Zerong Zheng and Chao Liang},
  title     = {{OmniHuman-1}: Rethinking the Scaling-Up of One-Stage Conditioned Human Animation Models},
  journal   = {arXiv preprint arXiv:2502.01061},
  year      = {2025}
}

@inproceedings{cui2025hallo3,
  title={Hallo3: Highly dynamic and realistic portrait image animation with video diffusion transformer},
  author={Cui, Jiahao and Li, Hui and Zhan, Yun and Shang, Hanlin and Cheng, Kaihui and Ma, Yuqi and Mu, Shan and Zhou, Hang and Wang, Jingdong and Zhu, Siyu},
  booktitle={Proceedings of the Computer Vision and Pattern Recognition Conference},
  pages={21086--21095},
  year={2025}
}

@article{jiang2025megatts,
  title={MegaTTS 3: Sparse Alignment Enhanced Latent Diffusion Transformer for Zero-Shot Speech Synthesis},
  author={Jiang, Ziyue and Ren, Yi and Li, Ruiqi and Ji, Shengpeng and Zhang, Boyang and Ye, Zhenhui and Zhang, Chen and Jionghao, Bai and Yang, Xiaoda and Zuo, Jialong and others},
  journal={arXiv preprint arXiv:2502.18924},
  year={2025}
}

@inproceedings{le2023voicebox,
  author    = {Matthew Le and Apoorv Vyas and Bowen Shi and Brian Karrer and Leda Sar{\i} and Rashel Moritz and Mary Williamson and Vimal Manohar and Yossi Adi and Jay Mahadeokar and Wei-Ning Hsu},
  title     = {{Voicebox}: Text-Guided Multilingual Universal Speech Generation at Scale},
  booktitle = neurips,
  year      = {2023}
}

@inproceedings{rombach2022high,
  title={{High-resolution image synthesis with latent diffusion models}},
  author={Rombach, Robin and Blattmann, Andreas and Lorenz, Dominik and Esser, Patrick and Ommer, Bj{\"o}rn},
  booktitle=cvpr,
  year={2022}
}

@inproceedings{ho2020denoising,
  title={{Denoising diffusion probabilistic models}},
  author={Ho, Jonathan and Jain, Ajay and Abbeel, Pieter},
  booktitle=neurips,
  year={2020}
}

@inproceedings{song2021scorebased,
    title={{Score-Based Generative Modeling through Stochastic Differential Equations}},
    author={Yang Song and Jascha Sohl-Dickstein and Diederik P Kingma and Abhishek Kumar and Stefano Ermon and Ben Poole},
    booktitle=iclr,
    year={2021},
}

@article{goodfellow2014generative,
  title={Generative adversarial nets},
  author={Goodfellow, Ian J and Pouget-Abadie, Jean and Mirza, Mehdi and Xu, Bing and Warde-Farley, David and Ozair, Sherjil and Courville, Aaron and Bengio, Yoshua},
  journal={Advances in neural information processing systems},
  volume={27},
  year={2014}
}

@inproceedings{yu2023celebv,
    title={{Celebv-text: A large-scale facial text-video dataset}},
    author={Yu, Jianhui and Zhu, Hao and Jiang, Liming and Loy, Chen Change and Cai, Weidong and Wu, Wayne},
    booktitle=cvpr,
    year={2023}
}

@inproceedings{wang2021one,
    title={{One-shot free-view neural talking-head synthesis for video conferencing}},
    author={Wang, Ting-Chun and Mallya, Arun and Liu, Ming-Yu},
    booktitle=cvpr,
    year={2021}
}

@inproceedings{esser2024scaling,
  title={{Scaling rectified flow transformers for high-resolution image synthesis}},
  author={Esser, Patrick and Kulal, Sumith and Blattmann, Andreas and Entezari, Rahim and M{\"u}ller, Jonas and Saini, Harry and Levi, Yam and Lorenz, Dominik and Sauer, Axel and Boesel, Frederic and others},
  booktitle=icml,
  year={2024}
}

@inproceedings{peebles2023scalable,
  title={{Scalable diffusion models with transformers}},
  author={Peebles, William and Xie, Saining},
  booktitle=iccv,
  year={2023}
}

@inproceedings{lipmanflow,
  title={{Flow Matching for Generative Modeling}},
  author={Lipman, Yaron and Chen, Ricky TQ and Ben-Hamu, Heli and Nickel, Maximilian and Le, Matthew},
  booktitle=iclr,
  year={2023}
}

@article{liuflow,
  title={Flow straight and fast: Learning to generate and transfer data with rectified flow},
  author={Liu, Xingchao and Gong, Chengyue and Liu, Qiang},
  journal={arXiv preprint arXiv:2209.03003},
  year={2022}
}

@misc{flux2024,
  author = {Black Forest Labs},
  title = {Flux.1},
  year = {2024},
  howpublished = {\url{https://blackforestlabs.ai/announcing-black-forest-labs/}},
  note = {Accessed: November 2024}
}

@article{yin2024improved,
  title={{Improved Distribution Matching Distillation for Fast Image Synthesis}},
  author={Yin, Tianwei and Gharbi, Micha{\"e}l and Park, Taesung and Zhang, Richard and Shechtman, Eli and Durand, Fredo and Freeman, William T},
  journal={arXiv preprint arXiv:2405.14867},
  year={2024}
}

@inproceedings{yin2024one,
  title={One-step diffusion with distribution matching distillation},
  author={Yin, Tianwei and Gharbi, Micha{\"e}l and Zhang, Richard and Shechtman, Eli and Durand, Fredo and Freeman, William T and Park, Taesung},
  booktitle=cvpr,
  year={2024}
}

@article{su2024roformer,
  title={Roformer: Enhanced transformer with rotary position embedding},
  author={Su, Jianlin and Ahmed, Murtadha and Lu, Yu and Pan, Shengfeng and Bo, Wen and Liu, Yunfeng},
  journal={Neurocomputing},
  volume={568},
  pages={127063},
  year={2024},
  publisher={Elsevier}
}

@article{saon17_interspeech,
  title={English conversational telephone speech recognition by humans and machines},
  author={Saon, George and Kurata, Gakuto and Sercu, Tom and Audhkhasi, Kartik and Thomas, Samuel and Dimitriadis, Dimitrios and Cui, Xiaodong and Ramabhadran, Bhuvana and Picheny, Michael and Lim, Lynn-Li and others},
  journal={arXiv preprint arXiv:1703.02136},
  year={2017}
}

@article{gao2025wan,
  title={Wan-s2v: Audio-driven cinematic video generation},
  author={Gao, Xin and Hu, Li and Hu, Siqi and Huang, Mingyang and Ji, Chaonan and Meng, Dechao and Qi, Jinwei and Qiao, Penchong and Shen, Zhen and Song, Yafei and others},
  journal={arXiv preprint arXiv:2508.18621},
  year={2025}
}

@article{ma2025playmate2,
  title={Playmate2: Training-Free Multi-Character Audio-Driven Animation via Diffusion Transformer with Reward Feedback},
  author={Ma, Xingpei and Huang, Shenneng and Cai, Jiaran and Guan, Yuansheng and Zheng, Shen and Zhao, Hanfeng and Zhang, Qiang and Zhang, Shunsi},
  journal={arXiv preprint arXiv:2510.12089},
  year={2025}
}

@article{chen2025humo,
  title={Humo: Human-centric video generation via collaborative multi-modal conditioning},
  author={Chen, Liyang and Ma, Tianxiang and Liu, Jiawei and Li, Bingchuan and Chen, Zhuowei and Liu, Lijie and He, Xu and Li, Gen and He, Qian and Wu, Zhiyong},
  journal={arXiv preprint arXiv:2509.08519},
  year={2025}
}

@article{tu2025stableavatar,
  title={Stableavatar: Infinite-length audio-driven avatar video generation},
  author={Tu, Shuyuan and Pan, Yueming and Huang, Yinming and Han, Xintong and Xing, Zhen and Dai, Qi and Luo, Chong and Wu, Zuxuan and Jiang, Yu-Gang},
  journal={arXiv preprint arXiv:2508.08248},
  year={2025}
}

@article{gan2025omniavatar,
  title={Omniavatar: Efficient audio-driven avatar video generation with adaptive body animation},
  author={Gan, Qijun and Yang, Ruizi and Zhu, Jianke and Xue, Shaofei and Hoi, Steven},
  journal={arXiv preprint arXiv:2506.18866},
  year={2025}
}

@article{chen2025hunyuanvideo,
  title={Hunyuanvideo-avatar: High-fidelity audio-driven human animation for multiple characters},
  author={Chen, Yi and Liang, Sen and Zhou, Zixiang and Huang, Ziyao and Ma, Yifeng and Tang, Junshu and Lin, Qin and Zhou, Yuan and Lu, Qinglin},
  journal={arXiv preprint arXiv:2505.20156},
  year={2025}
}

@article{yang2025infinitetalk,
  title={Infinitetalk: Audio-driven video generation for sparse-frame video dubbing},
  author={Yang, Shaoshu and Kong, Zhe and Gao, Feng and Cheng, Meng and Liu, Xiangyu and Zhang, Yong and Kang, Zhuoliang and Luo, Wenhan and Cai, Xunliang and He, Ran and others},
  journal={arXiv preprint arXiv:2508.14033},
  year={2025}
}

@article{wang2025talkverse,
  title={TalkVerse: Democratizing Minute-Long Audio-Driven Video Generation},
  author={Wang, Zhenzhi and Wang, Jian and Ma, Ke and Lin, Dahua and Zhou, Bing},
  journal={arXiv preprint arXiv:2512.14938},
  year={2025}
}

@article{seo2025lookahead,
  title={Lookahead anchoring: Preserving character identity in audio-driven human animation},
  author={Seo, Junyoung and Mira, Rodrigo and Haliassos, Alexandros and Bounareli, Stella and Chen, Honglie and Tran, Linh and Kim, Seungryong and Landgraf, Zoe and Shen, Jie},
  journal={arXiv preprint arXiv:2510.23581},
  year={2025}
}

@article{ding2025kling,
  title={Kling-avatar: Grounding multimodal instructions for cascaded long-duration avatar animation synthesis},
  author={Ding, Yikang and Liu, Jiwen and Zhang, Wenyuan and Wang, Zekun and Hu, Wentao and Cui, Liyuan and Lao, Mingming and Shao, Yingchao and Liu, Hui and Li, Xiaohan and others},
  journal={arXiv preprint arXiv:2509.09595},
  year={2025}
}

@article{li2025infinityhuman,
  title={InfinityHuman: Towards Long-Term Audio-Driven Human},
  author={Li, Xiaodi and Xie, Pan and Ren, Yi and Gan, Qijun and Zhang, Chen and Kong, Fangyuan and Yin, Xiang and Peng, Bingyue and Yuan, Zehuan},
  journal={arXiv preprint arXiv:2508.20210},
  year={2025}
}

@article{jiang2025omnihuman,
  title={Omnihuman-1.5: Instilling an active mind in avatars via cognitive simulation},
  author={Jiang, Jianwen and Zeng, Weihong and Zheng, Zerong and Yang, Jiaqi and Liang, Chao and Liao, Wang and Liang, Han and Zhang, Yuan and Gao, Mingyuan},
  journal={arXiv preprint arXiv:2508.19209},
  year={2025}
}

@inproceedings{guan2025audcast,
  title={Audcast: Audio-driven human video generation by cascaded diffusion transformers},
  author={Guan, Jiazhi and Wang, Kaisiyuan and Xu, Zhiliang and Yang, Quanwei and Sun, Yasheng and He, Shengyi and Liang, Borong and Cao, Yukang and Li, Yingying and Feng, Haocheng and others},
  booktitle={Proceedings of the IEEE/CVF Conference on Computer Vision and Pattern Recognition},
  pages={10678--10689},
  year={2025}
}

@article{zhang2025soul,
  title={Soul: Breathe Life into Digital Human for High-fidelity Long-term Multimodal Animation},
  author={Zhang, Jiangning and Zhu, Junwei and Gan, Zhenye and Luo, Donghao and Lin, Chuming and Xu, Feifan and Peng, Xu and Hu, Jianlong and Liu, Yuansen and Hong, Yijia and others},
  journal={arXiv preprint arXiv:2512.13495},
  year={2025}
}

@article{wang2026joyavatar,
  title={JoyAvatar: Unlocking Highly Expressive Avatars via Harmonized Text-Audio Conditioning},
  author={Wang, Ruikui and Feng, Jinheng and Tian, Lang and Luo, Huaishao and Li, Chaochao and Zhou, Liangbo and Zhang, Huan and Wu, Youzheng and He, Xiaodong},
  journal={arXiv preprint arXiv:2602.00702},
  year={2026}
}

@article{hacohen2026ltx,
  title={LTX-2: Efficient Joint Audio-Visual Foundation Model},
  author={HaCohen, Yoav and Brazowski, Benny and Chiprut, Nisan and Bitterman, Yaki and Kvochko, Andrew and Berkowitz, Avishai and Shalem, Daniel and Lifschitz, Daphna and Moshe, Dudu and Porat, Eitan and others},
  journal={arXiv preprint arXiv:2601.03233},
  year={2026}
}

@misc{veo3_2026,
  title        = {Veo 3},
  author       = {{Google DeepMind}},
  year         = {2026},
  howpublished = {\url{https://deepmind.google/models/veo/}},
  note         = {Accessed: 2026-03-01}
}

@article{wang2026klear,
  title={Klear: Unified Multi-Task Audio-Video Joint Generation},
  author={Wang, Jun and Qiang, Chunyu and Guo, Yuxin and Wang, Yiran and Zeng, Xijuan and Zhang, Chen and Wan, Pengfei},
  journal={arXiv preprint arXiv:2601.04151},
  year={2026}
}

@article{spleeter2020,
  title={Spleeter: a fast and efficient music source separation tool with pre-trained models},
  author={Hennequin, Romain and Khlif, Anis and Voituret, Felix and Moussallam, Manuel},
  journal={Journal of Open Source Software},
  volume={5},
  number={50},
  pages={2154},
  year={2020}
}

@article{huang2025self,
  title={Self forcing: Bridging the train-test gap in autoregressive video diffusion},
  author={Huang, Xun and Li, Zhengqi and He, Guande and Zhou, Mingyuan and Shechtman, Eli},
  journal={arXiv preprint arXiv:2506.08009},
  year={2025}
}

@article{shin2025motionstream,
  title={Motionstream: Real-time video generation with interactive motion controls},
  author={Shin, Joonghyuk and Li, Zhengqi and Zhang, Richard and Zhu, Jun-Yan and Park, Jaesik and Shechtman, Eli and Huang, Xun},
  journal={arXiv preprint arXiv:2511.01266},
  year={2025}
}
\newpage
\appendix
\setcounter{table}{0}
\setcounter{page}{1}
\setcounter{figure}{0}
\renewcommand{\thefigure}{A\arabic{figure}}
\renewcommand{\thetable}{A\arabic{table}}

\section*{Supplementary Material Overview}
This supplementary material provides additional implementation details, qualitative results, and analysis for JAM-Flow. We summarize the two-stage training pipeline, key architectural choices, and exclusive capabilities enabled by unified audio-motion generation. We also present further discussion of TTS performance and practical qualitative observations beyond the main paper. We strongly encourage reviewers to inspect the accompanying \textbf{HTML videos}: the qualitative results are \textbf{not cherry-picked}, and we intentionally include both successful cases and representative failure cases to provide a more faithful and transparent assessment of the method.

\section{Experimental Details}
\subsection{Dataset Preparation}
\paragraph{Data Sources.}
All of our models, including the first-stage Motion-DiT, were trained exclusively on the CelebV-Dub~\cite{sung2025voicecraft} dataset, which is derived from CelebV-HQ~\cite{zhu2022celebv} and CelebV-Text~\cite{yu2023celebv}. Both CelebV-Text and CelebV-HQ were sourced from the internet by their respective authors and are solely available for non-commercial research purposes. Videos are up to 30 seconds in length (the average is much shorter). For evaluation, clips longer than 20 seconds are segmented into non-overlapping 20-second chunks. Unless otherwise noted, we evaluate on the CelebV-Dub test split and HDTF~\cite{zhang2021flow}.

\paragraph{Audio and Motion Representations.}
For audio, we extract 100-dimensional mel-spectrograms with FFT size 1024, hop length 256, window size 1024, and a target sampling rate of 24 kHz. Facial motion is modeled at 25 FPS using preprocessed keypoint trajectories. We also define \emph{rest\_motion} ($\mathbf{e}^{\text{rest}}$) as the set of facial keypoints excluding the mouth region.

\paragraph{Known data limitations.}
CelebV-Dub contains pseudo-captions generated by Whisper and demuxed audio obtained via Spleeter~\cite{spleeter2020}. We observed non-trivial transcript errors and occasional artifacts in the demuxed audio. Further quantitative analysis and examples are provided in Sec.~\ref{sec:sup_dataset_considerations}.

\subsection{Model Architecture}
\paragraph{Backbones.}
As mentioned in the main paper, our Motion-DiT was trained from scratch using the CelebV-Dub dataset, while our Audio-DiT was initialized from the pre-trained F5-TTS model. To effectively harmonize these two models, we intentionally designed our Motion-DiT to have an architecture similar to that of Audio-DiT (\textit{i.e.}, F5-TTS). Both Audio-DiT and Motion-DiT share the same transformer backbone with 22 layers, a hidden dimension of 1024, and 16 attention heads (64 dimensions per head).

\paragraph{Joint attention layers.}
We insert $N_{\text{joint}}$ joint-attention blocks to couple the two streams. Unless otherwise specified, we adopt a \emph{Half Joint} configuration that fuses the early portion of the network (first half of the 22 layers) and keeps later layers modality-specific. This balances cross-modal fusion and modality-specialized capacity with substantially lower compute than \emph{Full Joint}.

We provide the implementations in Algorithm~\ref{alg:joint_dit_block} and Algorithm~\ref{alg:joint_attention}. To support modality-specific masking and enable configurations where audio joint attention can be disabled (\textit{e.g.}, for classifier-free guidance), we implemented the joint attention mechanism as shown in Algorithm~\ref{alg:joint_attention}. During training, all conditioning inputs (text, audio, and motion) are randomly dropped following the standard CFM training strategy. Rotary positional embeddings (RoPE) are scaled using a shared reference length to ensure temporal alignment across modalities. The code will be publicly released to support reproducibility and further research.

\subsection{Training and Inference Details}
\paragraph{Stage 1 Unimodal Training Details.}
Motion-DiT is trained from scratch using facial keypoint inputs, as visualized in Fig.~\ref{fig:liveportrait_vis}. The model is trained to predict clean mouth keypoints conditioned on audio and masked keypoint sequences. During this stage, we follow the standard talking head practice of concatenating wav2vec2 features with the motion inputs. These features are not used in Stage 2, where cross-modal conditioning is handled through joint attention. We directly adopt the pre-trained F5-TTS as our Audio-DiT.

\paragraph{Stage 2 Joint Training Details.}
Here, we connect the two distinct DiTs by adding joint-attention layers and applying RoPE alignment. Text conditioning is injected through the Audio-DiT branch, while the Motion-DiT branch receives \emph{rest\_motion} ($\mathbf{e}^{\text{rest}}$) and intermediate audio features ($\mathbf{f}^{\text{audio}}$) from the Audio-DiT via joint attention. Gradients flow across modalities within the joint blocks, enabling mutual refinement.

\paragraph{Condition dropping and Classifier-Free Guidance (CFG).}
To encourage robustness and enable a conditional sampling process, we apply modality-specific dropouts during training. Audio, \emph{(mouth) motion}, text, and \emph{rest\_motion} are dropped with probabilities of 0.1, 0.1, 0.2, and 0.8, respectively (i.e.,
\texttt{audio\_drop\_prob}=0.1,
\texttt{motion\_drop\_prob}=0.1,
\texttt{text\_drop\_prob}=0.2, 
\texttt{rest\_motion\_drop\_prob}=0.8).

These drops randomly mask conditioning signals, matching the inpainting-style supervision used throughout training. During inference, we optionally apply classifier-free guidance using a single shared guidance scale $\gamma$ for both motion and audio branches. In our setup, $\gamma$ mainly controls the overall strength of conditioning, and we find that a single shared value already works well in practice. We use $\gamma = 2$ for all reported results as a simple default, although we do not perform a dedicated sweep and better values may exist. Notably, the model remains strong even without CFG ($\gamma = 0$), suggesting that it already learns robust cross-modal representations.
\begin{equation}
    \hat{v}_\theta^{\text{motion}} = v_\theta^{\text{cond, motion}} + \gamma \cdot \left( v_\theta^{\text{cond, motion}} - v_\theta^{\text{uncond, motion}} \right)
\end{equation}

\begin{equation}
\hat{v}_\theta^{\text{audio}} = v_\theta^{\text{cond, audio}} + \gamma \cdot \left( v_\theta^{\text{cond, audio}} - v_\theta^{\text{uncond, audio}} \right)
\end{equation}

\paragraph{Training Compute.}
All training runs were conducted on four NVIDIA RTX 6000 Ada GPUs. Stage 1 and Stage 2 each take about one day.

\paragraph{Inference Speed.}
Our method's compact representation space enables remarkably fast inference. We used default settings for all baseline methods and 32 NFE (Number of Function Evaluations) for our method, as specified in the main paper. On a single RTX A6000 GPU, generating a 20-second lip-synced HDTF sample takes:
\begin{itemize}
  \item {Our method}: 45 seconds  / $\sim$500M parameters (joint audio + motion)
  \item {SadTalker}: 2.5 minutes (with GFPGAN)  / $\sim$200M parameters
  \item {AniPortrait}: 7 minutes / $\sim$1B parameters
  \item {Hallo}: 23 minutes / $1\sim3$B parameters
  \item {Hallo3}: 30 minutes (on an H100 GPU) / $1\sim3$B parameters
\end{itemize}
We note that Hallo3 requires GPUs with $>$48GB of memory and was tested on an H100 GPU (which is known to be several times faster than the A6000). Our training was conducted on 4 RTX 6000 Ada GPUs, while inference speed was measured using a single RTX A6000 GPU, except for Hallo3. As mentioned in the main paper, our method achieves real-time throughput of 25 fps on H100-class GPUs, taking 19 seconds to generate a 20-second video.
These results demonstrate a notable speedup over existing methods, making our approach practical for real-world applications while maintaining superior quality. Furthermore, our method could potentially be combined with diffusion distillation techniques such as DMD~\cite{yin2024improved, yin2024one, huang2025self, shin2025motionstream} to further reduce inference cost, potentially enabling real-time deployment on consumer-grade or even mobile devices.

\section{Qualitative Comparisons and  User Study}
We provide extensive qualitative comparisons across five categories in the supplementary webpage to complement the quantitative results in the main paper. We first discuss three standard tasks (talking head generation, text-to-speech, and automated video dubbing), and then present several additional inference cases enabled by our unique joint training. We also present user study results for two standard tasks: audio-conditioned talking head generation and automated video dubbing.

\subsection{Standard Tasks}

\begin{enumerate}
    \item Talking Head Generation: We present 14 test samples from the HDTF~\cite{zhang2021flow} dataset comparing our method (both I2V and V2V variants) against SadTalker~\cite{zhang2023sadtalker}, AniPortrait~\cite{wei2024aniportrait}, Hallo~\cite{xu2024hallo}, and Hallo3~\cite{cui2025hallo3}. Our V2V variant demonstrates superior lip-sync accuracy and more natural facial dynamics, achieving noticeably better synchronization with the input audio. Importantly, because our method uses keypoints beyond the mouth region as conditioning signals, it preserves the source motion and gestures more faithfully, producing outputs that are both lip-synchronized and motion-consistent.
    
    \item Text-to-Speech: We present 10 test samples from the LibriSpeech-PC test-clean set~\cite{panayotov2015librispeech, chen2024f5}, comparing our method with the F5-TTS baseline. Although standalone performance trails specialized TTS models due to pseudo-caption training, our audio quality remains strong for diverse creative applications, enabling audio-visual generation and cross-modal conditioning beyond the capabilities of pure TTS systems. While our model achieves a WER of 4.9\%, it effectively preserves speaker identity and produces speech that listeners judge to be natural and consistent. We also report results from Ours$^{\dagger}$, where the Audio-DiT is completely frozen, essentially replicating F5-TTS performance. This comparison highlights that our unified model, despite not being designed specifically as a TTS system, remains competitive in terms of intelligibility and voice similarity.

    \item Automated Video Dubbing: We present 15 test samples from CelebV-Dub~\cite{sung2025voicecraft}, comparing our method with HPMDubbing~\cite{cong2023learning}, StyleDubber~\cite{cong2024styledubber}, and VoiceCraft-Dub~\cite{sung2025voicecraft}. Our method produces synchronized audio-visual outputs without explicit task-specific optimization. Quantitative metrics such as LSE-C and LSE-D are unreliable in this setting (see the discussion in the main paper), making qualitative evaluation particularly important. The test videos were randomly selected; although our model sometimes produces subtle errors when matching prompts, competing methods also fail under similar conditions. Overall, the qualitative results demonstrate clear advantages in temporal alignment and speaker similarity.
\end{enumerate}

\subsection{Exclusive Use Cases and Failure cases}
We showcase several unique generation capabilities that emerge from our joint audio–motion modeling framework. These include: (1) text-to-multimodal synthesis that generates both audio and motion from text alone, (2) voice-preserving multimodal generation using reference audio to maintain speaker identity, (3) motion-constrained audio synthesis in which frozen motion guides audio generation with different semantic content, and (4) motion-to-audio generation without any text cues, demonstrating the model's ability to infer plausible speech from visual patterns alone. These diverse conditioning scenarios highlight the flexibility and cross-modal understanding of our unified approach.

\begin{enumerate}
\item Text $\rightarrow$ Audio + Motion.  
With only text input, the model \emph{jointly} generates both speech and synchronized lip motion, even without reference audio. The generated voices are drawn from random speaker identities, yet remain coherent with the motion. This setting directly validates the primary objective of our training: simultaneous audio–motion generation.

\item Text + Reference Audio $\rightarrow$ Audio + Motion.  
This setting is similar to Case 1, but is additionally conditioned on a short reference audio clip. The model produces speech in the target voice while generating matching lip motion. This again reflects the central joint audio–motion training objective, showing that the system can both preserve voice identity and synchronize motion with new text.

\item Reference Motion + Target Text $\rightarrow$ Audio.  
Here, we fix the video while changing the text. Perfect lip–audio alignment is impossible in this case, since the motion cannot be altered. Nevertheless, the generated audio aligns its timing (e.g., mouth opening and closure) with the visible motion, revealing that motion cues are attended to during speech generation. This case provides experimental evidence that unmasked motion features influence audio generation.

\item Reference Motion $\rightarrow$ Audio (without text).  
In this extreme setting, only motion is provided while text input is dropped. The model still produces plausible, time-aligned speech, showing that motion alone can guide audio synthesis. This demonstrates that the model has implicitly learned a mapping between visual articulation and acoustic patterns, beyond text-based conditioning.
\end{enumerate}

\paragraph{Failure Cases (Limitations).} 
Current limitations include (1) synchronization failures under significant length mismatches between modalities, where minor discrepancies may be absorbed through natural interjections but severe misalignments break lip-sync coherence; and (2) degraded performance on non-realistic inputs such as flat cartoons or highly stylized artwork, where the LivePortrait base model may fail to detect reliable keypoints.


\subsection{User Study}
\label{appedix:user}
\begin{figure}[h]
    \centering
    \begin{minipage}{0.49\textwidth}
        \centering
        \includegraphics[width=\textwidth]{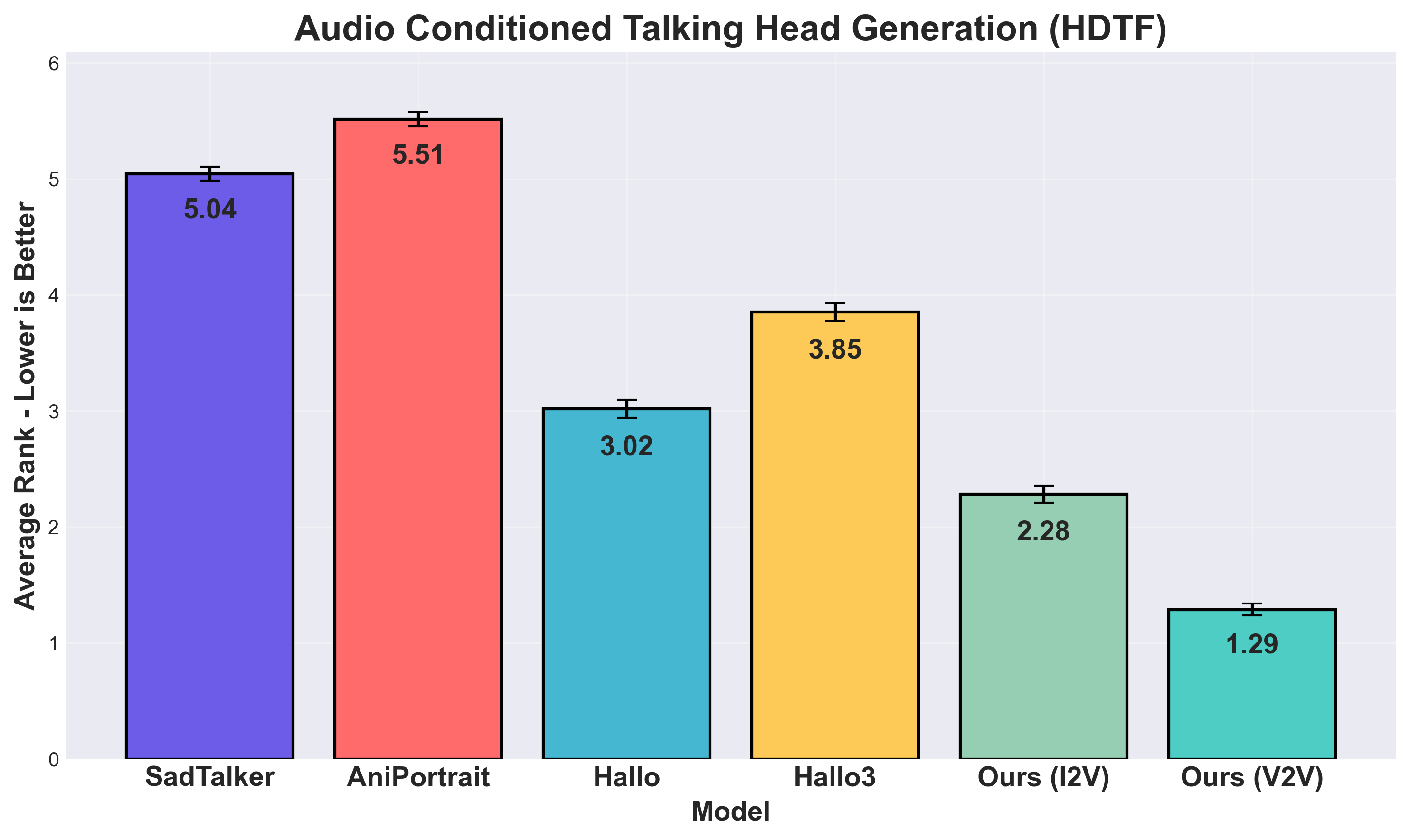}
        \caption{Average ranking results for audio-conditioned talking head generation on the HDTF dataset. Participants ranked six methods from best (1) to worst (6) based on overall quality, including lip-sync accuracy, motion naturalness, and visual fidelity. Lower ranks indicate better performance.}
        \label{fig:userstudy-th}
    \end{minipage}
    \hfill
    \begin{minipage}{0.49\textwidth}
        \centering
        \includegraphics[width=\textwidth]{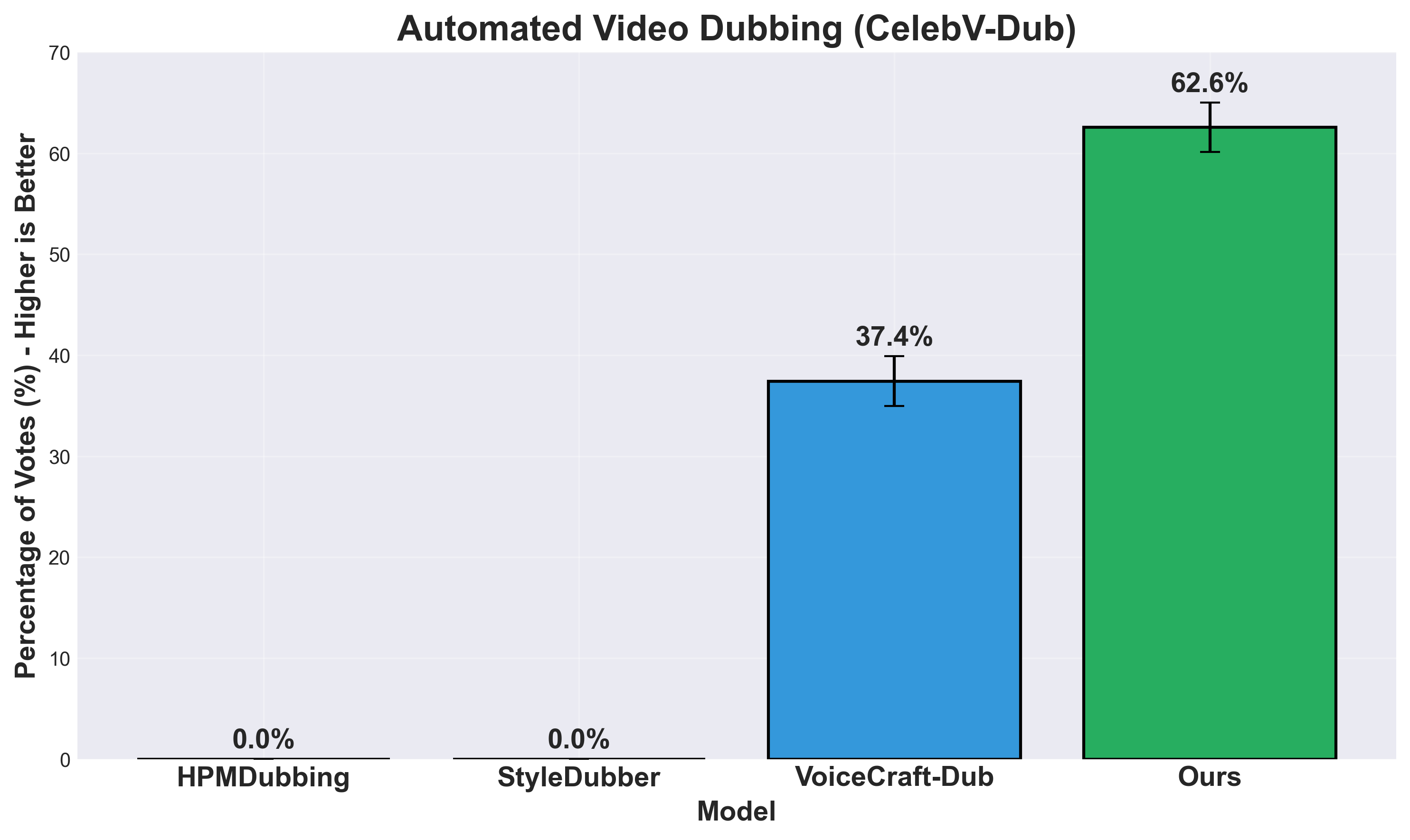}
        \caption{User preference results for automated video dubbing on the CelebV-Dub dataset. Participants selected the best synchronized audio-visual output among four competing methods for each sample. Values indicate the percentage of times each method was chosen as the best.}
        \label{fig:userstudy-dubbing}
    \end{minipage}
\end{figure}

We conducted a user study with 26 participants to evaluate perceptual quality across two tasks:
\begin{enumerate}
\item Audio-Conditioned Talking Head Generation (HDTF): Participants ranked six methods from best to worst for each sample. As shown in Figure~\ref{fig:userstudy-th}, our V2V variant achieved the best average rank of 1.29, followed by our I2V variant (2.28), significantly outperforming SadTalker (5.04), AniPortrait (5.51), Hallo (3.02), and Hallo3 (3.85).

\item Automated Video Dubbing (CelebV-Dub): Participants selected the best model among four methods for each sample. Figure~\ref{fig:userstudy-dubbing} shows that our method received 62.6\% of the votes, demonstrating a strong preference over VoiceCraft-Dub (37.4\%), while the other two methods received no votes.
\end{enumerate}

\begin{figure}[h]
  \centering
  \includegraphics[width=1.0\textwidth]{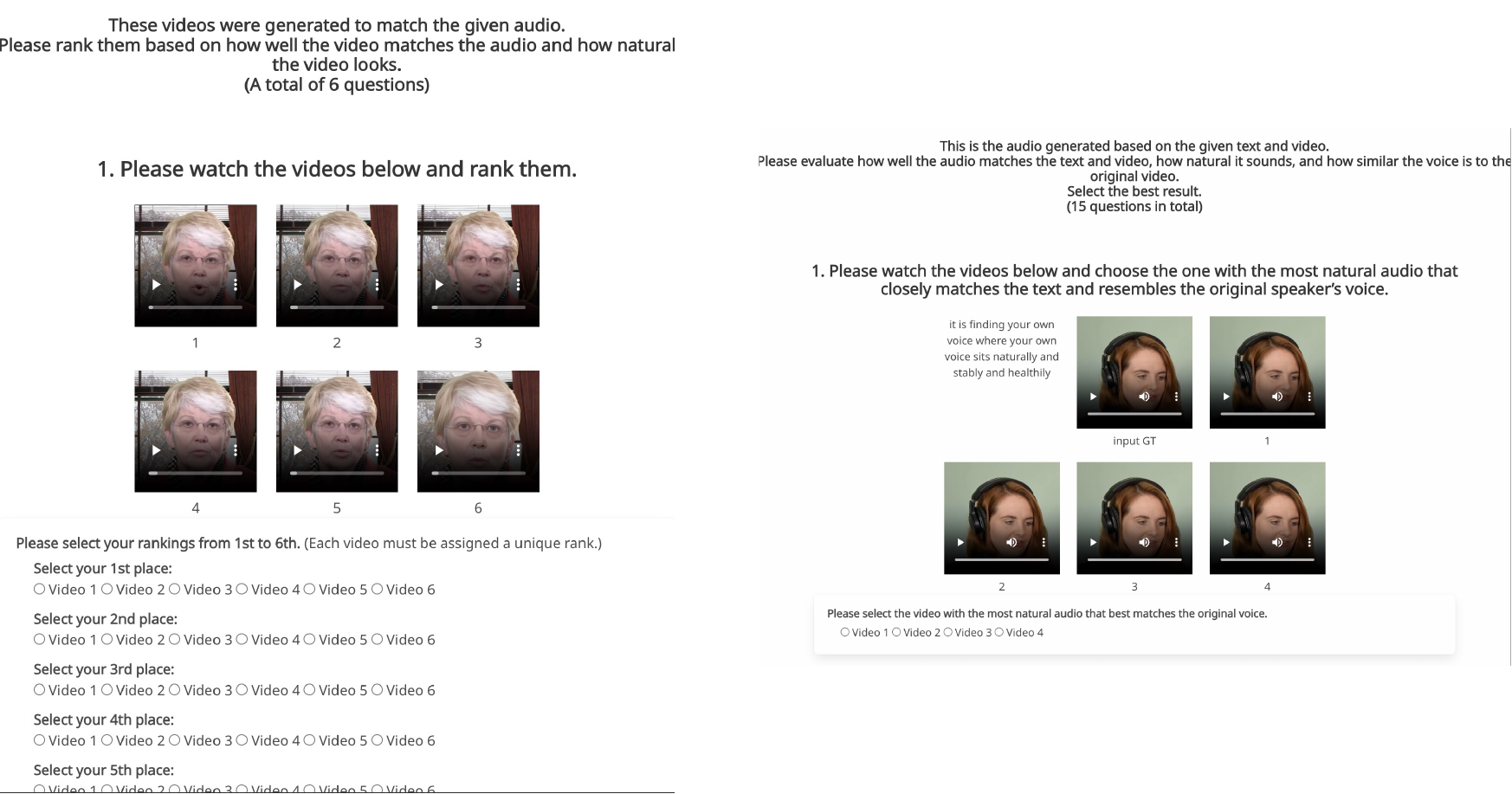}
  \caption{Survey examples. The left panel shows an example from the Audio-Conditioned Talking Head Generation (HDTF) survey, and the right panel shows an example from the Automated Video Dubbing (CelebV-Dub) survey.}
  \label{fig:survey_example}
\end{figure}

\section{Additional Discussions}
\subsection{LibriSpeech-PC Performance and Dataset Considerations}
Our model differs fundamentally from pure TTS systems in that it jointly optimizes for both audio and motion generation. Consequently, the observed WER gap (4.91\% vs.\ 2.42\%) reflects a multi-objective optimization trade-off. For Ours$^{\dagger}$, where the Audio-DiT is frozen, WER improves to 3.38\%, suggesting that part of the gap arises from dataset-related factors such as Whisper-generated captions and background-music removal.
A key challenge is the absence of publicly available datasets containing aligned video-audio-transcript triplets. CelebV-Dub~\cite{sung2025voicecraft} was therefore adopted as the only available proxy, constructed from CelebV-HQ~\cite{zhu2022celebv} and CelebV-Text~\cite{yu2023celebv} via background music removal and Whisper-based transcription. While this dataset enables joint training, it inevitably introduces transcription errors and audio artifacts.

\begin{itemize}
\item Transcript quality: We manually inspected all 213 test videos of CelebV-Dub, comparing audio against Whisper transcripts. Approximately 20\% of samples contained at least one incorrect word (45/213), and 30\% omitted the final word (72/213). The latter likely stems from Whisper’s difficulty in timestamping the last token. Hallucinations were also observed, \textit{e.g.}, ``They won't'' transcribed as ``They won't stop talking about it.''
\item Impact of background music removal: We evaluated whether the demuxing step itself degrades transcription. Applying the same background-music removal pipeline to clean F5-TTS outputs increased WER from 2.43\% to 2.63\%, suggesting that preprocessing artifacts may indeed contribute to the higher WER observed in CelebV-Dub.
\item Video modality noise: CelebV-Dub originates from YouTube videos recorded in uncontrolled environments, often with distant microphones, background noise, or sound effects. These factors introduce additional variability not present in curated audio-only corpora.
\end{itemize}

Taken together, these analyses suggest that the higher WER of our model is not solely a reflection of model quality, but also of dataset limitations. Our framework is not designed as a specialized TTS system; rather, it pioneers a unified formulation of joint speech–motion generation. Within this context, the performance observed on LibriSpeech-PC is consistent with both the multi-objective nature of our model and the imperfections of current pseudo-triplet datasets. Future work may benefit from more carefully curated datasets and advanced preprocessing pipelines, but our focus here is to demonstrate the feasibility and effectiveness of a unified architecture for multimodal co-generation.
\label{sec:sup_dataset_considerations}

\subsection{Qualitative Observations}
Across extensive experiments, we observed several consistent qualitative strengths. In talking head generation, JAM-Flow benefits from the compact LivePortrait keypoint representation, which enables fast inference while preserving identity and fine facial details. Compared with diffusion-based baselines such as Hallo and Hallo3, our outputs are typically more temporally stable, especially on longer sequences where competing methods often exhibit visible flicker.

In dubbing scenarios, the pretrained TTS prior helps produce speech that stays tightly aligned with lip motion while preserving natural pauses and coherent vocal tone. Notably, this audio-visual coherence is not explicitly supervised, suggesting that joint training captures cross-modal structure beyond simple temporal synchronization.

We also observe an asymmetric adaptation pattern during joint generation: when audio and motion conflict, the model tends to adjust the audio more than the visual motion. This bias preserves visual consistency while still achieving accurate lip-sync, and in practice leads to more natural results than cascaded or single-modal alternatives. Together with the user study results, these observations highlight the practical value of unified audio-motion modeling. We encourage readers to further inspect the ``Exclusive Use Cases'' and ``Failure Cases'' sections for representative examples and limitations.

\subsection{Usage of LLM and Reproducibility}
This paper only employed LLMs for polishing the text, not for generating content. The code and models will be released at a later stage to ensure reproducibility.

\begin{algorithm}[h]
\DontPrintSemicolon
\SetKwInput{KwIn}{Input}
\SetKwInput{KwOut}{Output}

\KwIn{
  $x_1,\,t_1$  — hidden state \& timestep for branch 1\;
  $x_2,\,t_2$  — hidden state \& timestep for branch 2\;
  
  $B_1=(\texttt{attn\_norm}_1,\texttt{attn}_1,\texttt{ff\_norm}_1,\texttt{ff}_1)$,\;
  $B_2=(\texttt{attn\_norm}_2,\texttt{attn}_2,\texttt{ff\_norm}_2,\texttt{ff}_2)$\;
  
  \texttt{mask}\(_1\), \texttt{mask}\(_2\) — optional local-window masks\;
  \texttt{rope}\(_1\), \texttt{rope}\(_2\) — rotary position embeddings\;
  
  $\alpha_1,\alpha_2\in\{0,1\}$ — flags: use \emph{joint} attention for each branch
}
\KwOut{$x_1',\,x_2'$ — updated hidden states}

\BlankLine
\textbf{/* 1. Normalization \& gating coefficients */}\;
$(\hat{x}_1,\;g^{\text{msa}}_1,\;s^{\text{ff}}_1,\;e^{\text{ff}}_1,\;g^{\text{ff}}_1)
 \leftarrow B_1.\texttt{attn\_norm}(x_1, t_1)$\;
$(\hat{x}_2,\;g^{\text{msa}}_2,\;s^{\text{ff}}_2,\;e^{\text{ff}}_2,\;g^{\text{ff}}_2)
 \leftarrow B_2.\texttt{attn\_norm}(x_2, t_2)$\;

\BlankLine
\textbf{/* 2. Joint or independent multi-head attention */}\;
$(a_1,\,a_2)\leftarrow
    \textsc{JointAttention}(B_1.\texttt{attn},\,B_2.\texttt{attn},
    \hat{x}_1,\hat{x}_2,\texttt{mask}_1,\texttt{mask}_2,
    \texttt{rope}_1,\texttt{rope}_2,\alpha_1,\alpha_2)$\;

\BlankLine
\textbf{/* 3. Residual connection with MSA gating */}\;
$x_1 \leftarrow x_1 + g^{\text{msa}}_1 \odot a_1$\;
$x_2 \leftarrow x_2 + g^{\text{msa}}_2 \odot a_2$\;

\BlankLine
\textbf{/* 4. Feed-forward branch — stream 1 */}\;
$\tilde{x}_1 \leftarrow
  B_1.\texttt{ff\_norm}(x_1)\odot(1+e^{\text{ff}}_1)
  +s^{\text{ff}}_1$\;
$f_1 \leftarrow B_1.\texttt{ff}(\tilde{x}_1)$\;
$x_1 \leftarrow x_1 + g^{\text{ff}}_1 \odot f_1$\;

\BlankLine
\textbf{/* 5. Feed-forward branch — stream 2 */}\;
$\tilde{x}_2 \leftarrow
  B_2.\texttt{ff\_norm}(x_2)\odot(1+e^{\text{ff}}_2)
  +s^{\text{ff}}_2$\;
$f_2 \leftarrow B_2.\texttt{ff}(\tilde{x}_2)$\;
$x_2 \leftarrow x_2 + g^{\text{ff}}_2 \odot f_2$\;

\BlankLine
\Return $(x_1,\,x_2)$\;
\caption{\textbf{JointDiTBlock} — single block of a diffusion transformer operating on two sequences with optional joint cross-attention. Element-wise multiplication is denoted by \(\odot\).}
\label{alg:joint_dit_block}
\end{algorithm}

\begin{algorithm}[b]
\DontPrintSemicolon
\SetKwInput{KwIn}{Input}
\SetKwInput{KwOut}{Output}

\KwIn{
  Attention modules $\texttt{attn}_1$, $\texttt{attn}_2$ with $QKV$ projections\;
  Hidden states $x_1 \in \mathbb{R}^{B \times L_1 \times d}$, $x_2 \in \mathbb{R}^{B \times L_2 \times d}$\;
  Optional rotary embeddings: \texttt{rope}$_1$, \texttt{rope}$_2$\;
  Optional local window masks: \texttt{mask}$_1$, \texttt{mask}$_2$\;
  Joint attention flags: $\alpha_1, \alpha_2 \in \{0,1\}$
}
\KwOut{
  Attention outputs $o_1 \in \mathbb{R}^{B \times L_1 \times d}$, $o_2 \in \mathbb{R}^{B \times L_2 \times d}$
}

\BlankLine
\textbf{1. Project input to QKV:} \\
$(q_1, k_1, v_1) \gets \texttt{attn}_1.\textsc{toQKV}(x_1)$ \\
$(q_2, k_2, v_2) \gets \texttt{attn}_2.\textsc{toQKV}(x_2)$

\BlankLine
\textbf{2. Apply rotary embeddings (if provided):} \\
\If{\texttt{rope}$_1$ exists}{
  $(q_1, k_1) \gets \textsc{ApplyRoPE}(q_1, k_1, \texttt{rope}_1)$
}
\If{\texttt{rope}$_2$ exists}{
  $(q_2, k_2) \gets \textsc{ApplyRoPE}(q_2, k_2, \texttt{rope}_2)$
}

\BlankLine
\textbf{3. Construct joint token pools:} \\
\eIf{$\alpha_1 = 1$}{
  $q_1^\star \gets [q_1; q_2]$,\quad $k_1^\star \gets [k_1; k_2]$,\quad $v_1^\star \gets [v_1; v_2]$
}{
  $(q_1^\star, k_1^\star, v_1^\star) \gets (q_1, k_1, v_1)$
}

\eIf{$\alpha_2 = 1$}{
  $q_2^\star \gets [q_2; q_1]$,\quad $k_2^\star \gets [k_2; k_1]$,\quad $v_2^\star \gets [v_2; v_1]$
}{
  $(q_2^\star, k_2^\star, v_2^\star) \gets (q_2, k_2, v_2)$
}

\BlankLine
\textbf{4. Split heads and apply masks:} \\
$(q_1^\star, k_1^\star, v_1^\star) \gets \textsc{SplitHeads}(q_1^\star, k_1^\star, v_1^\star)$ \\
$(q_2^\star, k_2^\star, v_2^\star) \gets \textsc{SplitHeads}(q_2^\star, k_2^\star, v_2^\star)$

\BlankLine
\If{$\texttt{mask}_1 \neq \varnothing \wedge \alpha_1 = 1$}{
  $M_1 \gets \textsc{CustomDiagMask}(L_1, L_2, \texttt{mask}_1)$
}
\Else{
  $M_1 \gets \varnothing$
}

\If{$\texttt{mask}_2 \neq \varnothing \wedge \alpha_2 = 1$}{
  $M_2 \gets \textsc{CustomDiagMask}(L_2, L_1, \texttt{mask}_2)$
}
\Else{
  $M_2 \gets \varnothing$
}

\BlankLine
\textbf{5. Compute scaled dot-product attention:} \\
$o_1^\star \gets \textsc{SDPA}(q_1^\star, k_1^\star, v_1^\star, M_1)$ \\
$o_2^\star \gets \textsc{SDPA}(q_2^\star, k_2^\star, v_2^\star, M_2)$

\BlankLine
\textbf{6. Merge heads, trim to original length, and project:} \\
$o_1 \gets \textsc{MergeHeads}(o_1^\star)[:, :L_1]$,\quad $o_1 \gets \texttt{attn}_1.\textsc{outProj}(o_1)$ \\
$o_2 \gets \textsc{MergeHeads}(o_2^\star)[:, :L_2]$,\quad $o_2 \gets \texttt{attn}_2.\textsc{outProj}(o_2)$

\BlankLine
\Return $(o_1, o_2)$

\caption{\textbf{JointAttention} – Multi-head attention with optional joint token pooling.  
Concatenation $[\,\cdot\,; \cdot\,]$ is along the sequence dimension.  
\textsc{SDPA} denotes scaled dot-product attention.  
\textsc{CustomDiagMask} is a full+diagonal window mask, illustrated in Fig.~\ref{fig:architecture}.}
\label{alg:joint_attention}
\end{algorithm}

\end{document}